\documentclass{article}
\usepackage[preprint]{neurips_2023}

\usepackage[utf8]{inputenc} % allow utf-8 input
\usepackage[T1]{fontenc}    % use 8-bit T1 fonts
\usepackage{hyperref}       % hyperlinks
\usepackage{url}            % simple URL typesetting
\usepackage{booktabs}       % professional-quality tables
\usepackage{amsfonts}       % blackboard math symbols
\usepackage{nicefrac}       % compact symbols for 1/2, etc.
\usepackage{microtype}      % microtypography
\usepackage{xcolor}         % colors

\usepackage{amsmath,amsfonts,bm}

\def\eqref#1{equation~\ref{#1}}
\def\1{\bm{1}}

\DeclareMathAlphabet{\mathsfit}{\encodingdefault}{\sfdefault}{m}{sl}
\SetMathAlphabet{\mathsfit}{bold}{\encodingdefault}{\sfdefault}{bx}{n}

\usepackage{amsmath}
\usepackage{amssymb}
\usepackage{mathtools}
\usepackage{amsthm}

\usepackage{multirow}
\usepackage{graphicx}
\usepackage{subcaption}
\usepackage{arydshln}
\usepackage{wrapfig}
\usepackage{enumitem}

\usepackage[capitalize,noabbrev]{cleveref}
\crefname{figure}{Fig.}{Fig.}
\crefname{section}{Sec.}{Sec.}
\crefname{table}{Table}{Tables}

\title{CodeT5+: Open Code Large Language Models for Code Understanding and Generation}

\author{%
  Yue Wang\thanks{Equal contribution. Corresponding authors: \texttt{\small \{wang.y, hungle, shoi\}@salesforce.com}} , Hung Le\footnotemark[1] ,  Akhilesh Deepak Gotmare, Nghi D.Q. Bui, Junnan Li, Steven C.H. Hoi
  \\
  Salesforce AI Research\\
  \url{https://github.com/salesforce/CodeT5/tree/main/CodeT5+} 
}  

\begin{document}

\maketitle

\begin{abstract}
Large language models (LLMs) pretrained on vast source code have achieved prominent progress in code intelligence. However, existing code LLMs have two main limitations in terms of architecture and pretraining tasks. First, they often adopt a specific architecture (encoder-only or decoder-only) or rely on a  unified encoder-decoder network  for different downstream tasks. The former paradigm is limited by inflexibility in applications while in the latter, the model is treated as a single system for all tasks, leading to suboptimal performance on a subset of tasks. Secondly, they often employ a limited set of pretraining objectives which might not be relevant to some downstream tasks and hence result in substantial performance degrade. To address these limitations, we propose “CodeT5+”, a family of encoder-decoder LLMs for code in which component modules can be flexibly combined to suit a wide range of downstream code tasks. Such flexibility is enabled by our proposed  mixture of pretraining objectives to mitigate the pretrain-finetune discrepancy. These objectives cover span denoising, contrastive learning, text-code matching, and causal LM pretraining tasks, on both unimodal and bimodal multilingual code corpora. Furthermore, we propose to initialize CodeT5+ with frozen off-the-shelf LLMs without training from scratch to efficiently scale up our models, and explore instruction-tuning to align with natural language instructions. We extensively evaluate CodeT5+ on over $20$ code-related benchmarks in different settings, including zero-shot, finetuning, and instruction-tuning. We observe state-of-the-art (SoTA) model performance on various code-related tasks, such as code generation and completion, math programming, and text-to-code retrieval tasks. Particularly, our instruction-tuned CodeT5+ 16B achieves new SoTA results of $35.0\%$ pass@1 and $54.5\%$ pass@10 on the HumanEval code generation task against other open code LLMs, even surpassing the OpenAI code-cushman-001 model.
\end{abstract}

\section{Introduction}
Large language models (LLMs)  \citep{codex, codet5, codegen} have recently demonstrated remarkable success in a broad set of downstream tasks in the code domain \citep{csn, codexglue, apps}.
By pretraining  on massive code-based data (e.g. GitHub public data), these code LLMs can learn rich contextual representations which can be transferred to various code-related downstream tasks. 
However, we found that many of the existing models are designed to perform well only in a subset of tasks. We argue that this is mainly due to two  limitations in terms of architecture and pretraining tasks.

From an architectural perspective, existing code LLMs often adopt encoder-only or decoder-only models that perform well only on certain understanding or generative tasks.
Specifically, encoder-only models \citep{codebert, graphcodebert} are often used to facilitate understanding tasks such as text-to-code retrieval \citep{codexglue}.
For generative tasks such as code generation \citep{codex, apps}, decoder-only models \citep{codex, codegen} often demonstrate stronger performance.
However, these decoder-only models are often not ideal for understanding tasks such as retrieval and detection tasks compared to encoder-only models \citep{codegen2}.
Besides, several recent models have adopted more unified encoder-decoder architectures ~\citep{codet5,plbart} to adapt to different types of tasks. 
While these models can support both understanding and generative tasks, they still suffer from suboptimal performance on certain tasks.
\citet{unixcoder} found that encoder-decoder models fail to beat state-of-the-art (SoTA) encoder-only or decoder-only baselines on retrieval and code completion tasks respectively.
This shortfall is due to the limitation of the single-module architecture generally adapted to all tasks. 
In summary, prior approaches are not designed with compositionality such that individual components can be activated to better suit different types of downstream tasks. 

From a learning objective perspective, current models employ a limited set of  pretraining tasks.
These tasks can lead to performance degrade on certain downstream tasks due to the  discrepancy between the pretraining and finetuning stage.
For instance, T5-based models such as \citep{codet5} are often trained with a span denoising objective. However, in downstream tasks such as code generation \citep{codex, apps}, most state-of-the-art models are pretrained with a next-token prediction objective which auto-regressively predicts a program token by token. 
Furthermore, many models are not trained to learn contrastive code representations that are vital for understanding tasks such as text-to-code retrieval. Although recent attempts~\citep{unixcoder,syncobert} introduce a contrastive learning task to alleviate this issue, these approaches ignore the fine-grained cross-modal alignments between text and code representations. 

\begin{figure*}[t]
	\centering
	\resizebox{1.0\textwidth}{!} {
 \includegraphics{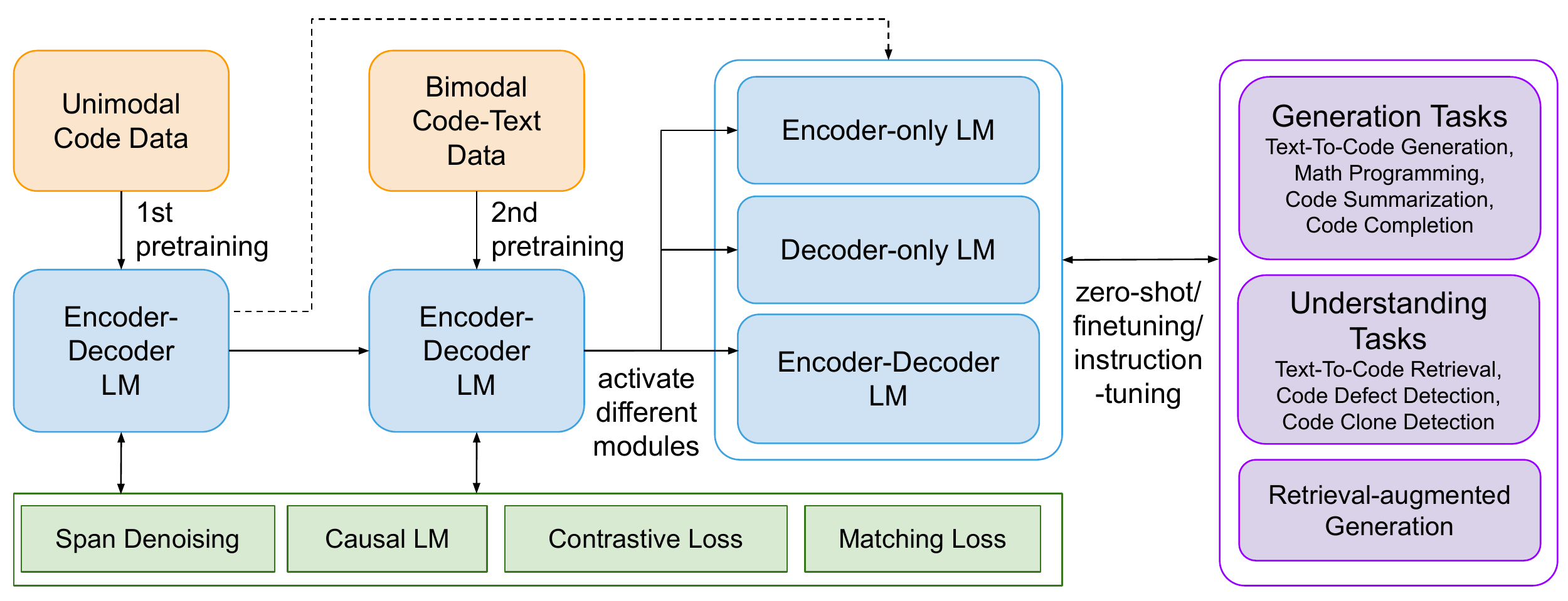}
	}
	\caption{
        \textbf{An overview of our CodeT5+ approach:}
        CodeT5+ is a family of code large language models to address a wide range of code understanding and generation tasks. The framework contains a diverse mixture of pretraining objectives on unimodal and bimodal data. 
        Individual modules of CodeT5+ can be flexibly detached and combined to suit different downstream applications in zero-shot, finetuning, or instruction-tuning settings.
        }
	\label{fig:intro_overview}
\end{figure*}

To address the above limitations, we propose ``CodeT5+'',
a new family of encoder-decoder code foundation   LLMs
for a wide range of code understanding and generation tasks (see \cref{fig:intro_overview} for an overview). 
Despite being an encoder-decoder based model, our CodeT5+ can flexibly operate in encoder-only, decoder-only, and encoder-decoder modes to suit different downstream applications. Such flexibility is enabled by our proposed pretraining tasks, which include span denoising and causal language modeling (CLM) tasks on code data and 
text-code contrastive learning, matching, and CLM tasks on text-code data.
We found that such a wide set of pretraining tasks can help learn rich representations from both code and text data, and bridge the pretrain-finetune gap in various applications. Besides, we show that the integration of the matching task with contrastive learning is crucial to capture the fine-grained text-code alignments and improve retrieval performance.

Furthermore, we scale up the model size of CodeT5+ with a compute-efficient pretraining strategy by leveraging off-the-shelf code LLMs~\citep{codegen} to initialize the  components of CodeT5+.
Specifically, we employ a ``shallow encoder and deep decoder'' architecture~\citep{alphacode}, where both encoder and decoder are initialized from pretrained checkpoints and  connected by cross-attention layers. We freeze the deep decoder LLM and only train the shallow encoder and cross-attention layers, largely reducing the number of trainable parameters for efficient tuning. 
Finally, recent work in the NLP domain \citep{alpaca, wang2022self, ouyang2022training} inspired us to explore CodeT5+ with instruction tuning to better align the models with natural language instructions. 

We extensively evaluate CodeT5+ on over $20$ code-related benchmarks under various settings, including zero-shot, finetuning, and instruction-tuning.
Results show that CodeT5+ yields substantial performance gains on many downstream tasks compared to their SoTA  baselines, e.g., $8$ text-to-code retrieval tasks ($+3.2$ avg. MRR), $2$ line-level code completion tasks ($+2.1$ avg. Exact Match), and $2$ retrieval-augmented code generation tasks ($+5.8$ avg. BLEU-4). 
In $2$ math programming tasks on MathQA and GSM8K benchmarks \citep{austin2021program, gsm}, CodeT5+ models of below billion-parameter sizes significantly outperform many LLMs of up to 137B parameters.
Particularly, in the zero-shot text-to-code generation task on HumanEval benchmark \citep{codex}, our instruction-tuned CodeT5+ 16B sets new SoTA results  of $35.0\%$ pass@1 and $54.5\%$ pass@10 against other open code LLMs, even surpassing the closed-source OpenAI code-cushman-001 model.
Finally,  we showcase that CodeT5+ can be seamlessly adopted as a semi-parametric {retrieval-augmented generation} system which significantly outperforms similar methods in code generation.
All CodeT5+ models will be open-sourced to support the research and developer communities.

\section{Related Work}

Following the success of large language models (LLMs) such as BERT~\citep{bert} and GPT~\citep{gpt} in natural language processing (NLP), recent years witness a surge of research work of LLMs in the code domain, leading to new SoTA results on a wide spectrum of code-related tasks. 
Typically, code-based  LLMs  can be  categorized into three architectures: encoder-only models \citep{codebert, graphcodebert, codemvp}, decoder-only models ~\citep{codexglue, codex, incoder, codegen}, and encoder-decoder models ~\citep{plbart, codet5, sptcode, natgen}.
For encoder-only and decoder-only models, they are often ideal for either understanding tasks such as code retrieval \citep{csn} or generation tasks such as code synthesis \citep{codex, apps} respectively. 
For encoder-decoder models, they can be adapted to both code understanding and generation but do not always achieve better performance \citep{codet5, plbart} than decoder-only or encoder-only models. 
In this work, we propose a new family of encoder-decoder code large language models  that can flexibly operate in various modes, including encoder-only, decoder-only, and encoder-decoder models.

Prior code LLMs are also limited by their pretraining tasks, which are not perfect to transfer the models to some downstream tasks.
For instance,  T5-based models such as \citep{codet5} pretrained with span denoising objective are not ideal for auto-regressive generation tasks like next-line code completion \citep{codexglue,gptc}, as these models are trained to recover short spans of limited lengths rather than a whole program.\footnote{{Recently, \citet{tabachnyk2022ml, incoder} demonstrated using encoder-decoder models for infilling-style code completion, in which code context after the cursor is provided. Such code completion setting is not our focus in this work.  
}} 
Inspired by recent advances in NLP research \citep{ul2,alexatm}, we explore to combine span denoising with CLM tasks to improve the model with better causal generation capability \citep{coderl}.
Additionally, most models do not have specific pretraining tasks (e.g. contrastive learning) to facilitate the learning of contextual representations that can distinguish code samples of different semantics. This can lead to suboptimal performance on code understanding tasks like code retrieval \citep{csn}.
In light of this observation, in our pretraining objectives, we include a contrastive learning task to learn better unimodal  representations and a matching task to learn richer bimodal representations.
These tasks have demonstrated positive impacts in related vision-language pretraining ~\citep{albef}.

More related to our work is UniXcoder~\citep{unixcoder}, which adopts a UniLM-style design ~\citep{unilm} and supports various tasks by manipulating input attention masks. 
However, as the model attempts to rely on a single encoder  to support all tasks, UniXcoder  suffers from the inter-task interference, leading to performance degrade especially on sequence-to-sequence tasks such as code generation.
UniXcoder and related work \citep{codet5, unixcoder, codemvp} also use code-specific features such as abstract syntax trees and identifiers.
In CodeT5+, we efficiently activate component modules for different tasks and do not rely on code-specific features. 

Finally, also related to our work is the research of parameter-efficient LLM training which aims to scale LLMs using limited computation resources.
A common strategy to achieve this goal is to only train a small number of (extra) model parameters while freezing a large part of LLM \citep{hu2022lora, Sung_2022_CVPR}.
Another common feature is the use of prompting, either with continuous or discrete prompts, to efficiently align models to downstream tasks \citep{liu2021gpt, lester-etal-2021-power, liu-etal-2022-p, ponti-etal-2023-combining}.
In this work, we scale our models by leveraging LLMs to initialize the encoder and decoder components of CodeT5+ with pretrained model checkpoints.
We employ a ``shallow encoder and deep decoder'' architecture  by \citet{alphacode} and only keep the small encoder and the cross-attention layers  trainable while freezing the deep decoder LLM. 
We then combine this training scheme with instruction tuning \citep{alpaca, wang2022self, ouyang2022training}, using only a small set of synthetic instruction-following prompts by~\citet{codealpaca}, to efficiently guide CodeT5+ towards better alignment to downstream tasks. 

\begin{figure*}[t]
	\centering
	\resizebox{1\textwidth}{!} {
 \includegraphics{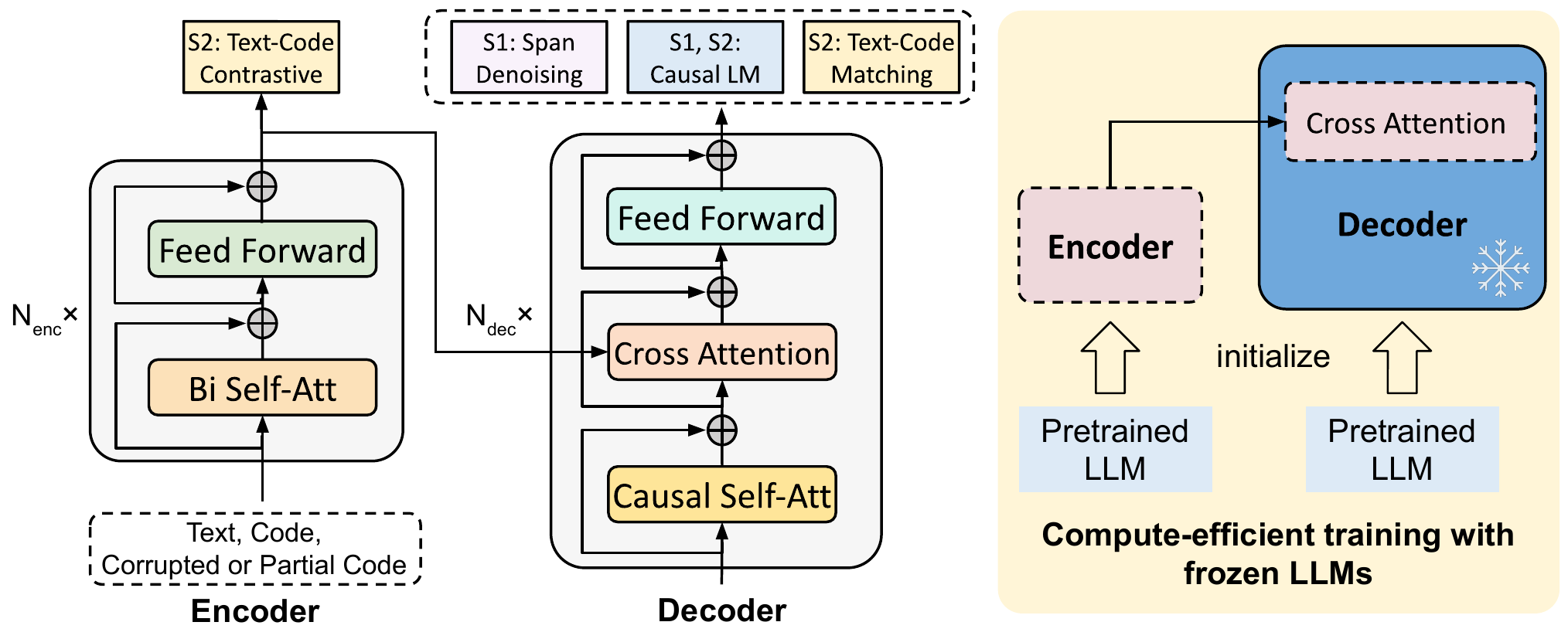}
	}
	\caption{
        \textbf{Model architecture:}
        The encoder learns to encode contextual representations from code/text sequences (either complete, partial, or span-masked sequences)
        while
        the decoder is trained to generate different types of outputs, depending on the pretraining learning tasks: 
        S1: first stage pretraining with unimodal code corpus.
        S2: second stage pretraining with bimodal code-text corpus. 
        The diagram on the right illustrates our proposed compute-efficient training with frozen code LLMs to scale up the model.
        We employ a ``shallow encoder and deep decoder'' architecture  and only keep the small encoder and the cross-attention layers  trainable while freezing the deep decoder LLM. 
        }
	\label{fig:architecture}
\end{figure*}

\section{CodeT5+: Open Code Large Language Models}
We develop CodeT5+, a new family of open code large language models for code understanding and generation tasks (see \cref{fig:intro_overview} for an overview and more architecture/pretraining details in  \cref{fig:architecture} and \cref{fig:pretrain_objectives}). 
Based on the encoder-decoder architecture ~\citep{codet5}, CodeT5+ is enhanced with the flexibility to operate in various modes for different downstream tasks through our proposed mixture of pretraining objectives on unimodal and bimodal data.

In the first stage of unimodal pretraining, we pretrain the model with massive code data using computationally efficient objectives (\cref{subsec:unimodal_pretrain}).
In the second stage of bimodal pretraining, we continue to pretrain the model with a smaller set of  code-text data with  cross-modal learning objectives (\cref{subsec:bimodal_pretrain}).
For each stage, we jointly optimize multiple pretraining objectives with equal weights. 
We found that this stage-wise training approach can efficiently expose our models to more diverse data to learn rich contextual representations.
Additionally, we explore initializing CodeT5+ with off-the-shelf code LLMs to efficiently scale up the model (\cref{subsec:integration}).
Finally, model components in CodeT5+ can be dynamically combined to suit different downstream application tasks (\cref{subsec:finetuning}).

\subsection{Unimodal Pretraining on Code Data}\label{subsec:unimodal_pretrain}
\begin{figure*}[t]
	\centering
	\resizebox{1.0\textwidth}{!} {
	\includegraphics{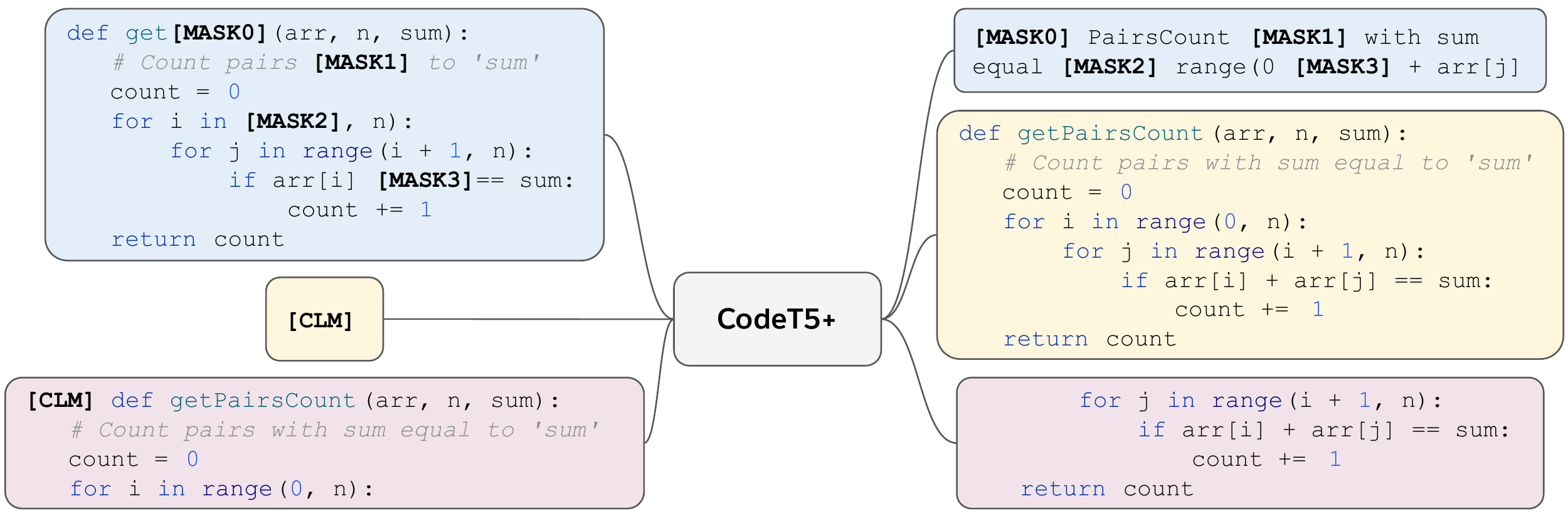}
	}
	\caption{
        \textbf{Self-supervised pretraining on open-source code data:} 
        we pretrain CodeT5+ on code data using a mixture of tasks: 
        (i) span denoising (Top);
        (ii) decoder-only causal LM (Middle); and 
        (iii) Seq2Seq causal LM (Bottom).
        This mixture of tasks lets the models learn meaningful representations of code contexts and recover missing information at different levels: code spans, partial programs, and complete programs.
        }
	\label{fig:pretrain_objectives}
\end{figure*}

In the first stage, we pretrain CodeT5+ on large-scale code unimodal data, which can be obtained from open-source platforms like GitHub. 
Although such data also contain texts such as user-written code comments, we denote unimodal data to distinguish them with bimodal data of text-code pairs in the second pretraining stage.
It is non-trivial to separate the code and text due to various commenting styles of programmers and different commenting syntax of languages. 
In this stage, we pretrain the model from scratch using a mixture of span denoising and CLM tasks as shown in \cref{fig:pretrain_objectives}.
These tasks enable the model to learn to recover code contexts at different scales: code spans, partial programs, and complete programs.

\textbf{Span Denoising.} Similar to T5~\citep{t5}, we randomly replace $15\%$  of the tokens with indexed sentinel tokens (like \texttt{[MASK0]}) in the encoder inputs, and require the decoder to recover them via generating a combination of these spans. 
We follow CodeT5 to employ whole-word masking by sampling spans 
(span lengths determined by a uniform distribution with a mean of $3$) 
before subword tokenization to avoid masking partial words. 
To accelerate the training, we concatenate different code files into sequences and truncate them into chunks of fixed length.

\textbf{Causal Language Modeling (CLM).}
Inspired by~\citet{ul2,alexatm}, we introduce two variants of CLM to optimize our model for auto-regressive generation.
In the first variant, we randomly select a pivot location and regard the context before it as the source sequence and the sequence after it as the target output.
We denote this variant as a sequence-to-sequence (Seq2Seq) causal LM objective.
We restrict the pivot location to be uniformly sampled between $10\%$ and $90\%$ of the whole sequence and  prepend a special token \texttt{[CLM]} to the source sequence. 
The second CLM variant is a decoder-only generation task and can be viewed as an extreme case of the first variant. 
In this task, we always pass a  \texttt{[CLM]} token to 
 the encoder input  and require the decoder to generate the full code sequence.
 Compared to the first variant, this task aims to provide more dense supervision signals to train the decoder as an independent full-fledged code generation module. 
 %For both CLM tasks, we only feed a single code file to the model instead of concatenating multiple files in the span denoising objective.

\subsection{Bimodal Pretraining on Text-code Data}\label{subsec:bimodal_pretrain}
In the second stage, we pretrain the model using text-code bimodal data at the function level ~\citep{csn}.
In this setting, each text-code pair contains a code function and its corresponding docstring describing its semantics. 
Such a bimodal data format facilitates model training for cross-modal understanding and generation. 
The bimodal pretraining tasks consist of cross-modal contrastive learning, matching, and causal LM tasks, as shown in \cref{fig:architecture}

\paragraph{Text-Code Contrastive Learning.} This task aims to align the feature space of text and code representations by pulling together the representations of positive text-code pairs and pulling apart the negative pairs.
\citet{unixcoder} demonstrated the benefits of such learning task for code understanding. 
This task only activates the encoder, which encodes a text or code snippet into a continuous representation through bidirectional self-attention \citep{transformer}. Similar to BERT~\citep{bert}, we prepend a special token \texttt{[CLS]} to the input and regard its output embeddings at the final Transformer layer as the representations of the corresponding input text or code.  We further add a linear layer and use L2 normalization to map the output to 256-dimensional embeddings. 
To enrich the negative samples, we use a momentum encoder to store embeddings of samples from previous mini-batches, as similarly adopted by~\cite{moco,blip}.
Specifically, the momentum encoder maintains a queuing system that enqueues the samples in the current mini-batch and dequeues the samples in the oldest mini-batch. 
We update the momentum encoder by linear interpolation of the original encoder and the momentum encoder to ensure the consistency of representations across training steps.

% To ensure the consistency of representations across training steps, we update the momentum encoder by linear interpolation of the original encoder and the momentum encoder.
% Besides, since text and code samples might be loosely paired and each text/code sample can have multiple positive pairs, we also use the momentum encoder to create soft labels and consider the potential positives in the negative pairs. 

\paragraph{Text-Code Matching.} 
This task activates the decoder and aims to predict whether a text and code snippet share the same semantics.
Such task enables model to learn better bimodal representations that capture the fine-grained alignment between text and code modalities. 
 Given a code sample, the decoder first passes it to an embedding layer and a causal self-attention layer. 
The self-attention representations are then passed to a cross-attention layer which queries relevant signals from the text representations (received from the encoder). 
A task-specific \texttt{[Match]} token is prepended to the code input sequence to inform the decoder of the text-code matching functionality, and an \texttt{[EOS]} token is appended to the end of the code input. 
 Since the decoder employs causal self-attention masks and only the last decoder token can attend to the whole context, we treat the output embedding of \texttt{[EOS]} at the last decoder layer as the text-code cross-modal alignment representation.
Finally, we use a linear layer on top of the output embedding of the decoder for a binary matching task, predicting whether a text-code pair is positive (matched) or negative (unmatched).

In order to find more informative negatives, we employ a hard negative mining strategy \citep{albef}.
 Specifically, we sample hard negatives based on the contrastive-based similarity scores between the current sample and previous samples in the queue maintained by the momentum encoder.
 As such, harder negatives are more likely to be selected. 
For a batch of positive pairs, we construct two batches of negative pairs by mining negatives from the text/code queue with a code/text query.

\paragraph{Text-Code Causal LM.} This task activates both encoder and decoder and focuses on a cross-modal generative objective through a dual multimodal conversion: text-to-code generation and code-to-text generation. 
Specifically, when the input is a text sample, we prepend a \texttt{[CDec]} token to the input sequence to the decoder.
In this case, the decoder operates under code generation functionality. 
Alternatively, when the input is a code sample, we  prepend a \texttt{[TDec]} token to the input sequence to the decoder.
The decoder operates under text generation functionality in this case.  
This type of Causal LM has been shown to be an effective learning objective to close the pretrain-finetune gap for multimodal generative downstream tasks such as code summarization \citep{codet5}.

\subsection{Compute-efficient Pretraining with Frozen Off-the-shelf LLMs}\label{subsec:integration}
To efficiently scale up the model without the need of pretraining from scratch, we  propose a compute-efficient pretraining strategy to initialize model components (i.e. encoder and decoder) of CodeT5+ with off-the-shelf pretrained LLMs~\citep{codegen} (see the rightmost diagram of \cref{fig:architecture}).
For this extension,  inspired by~\citep{alphacode},  we employ a ``shallow encoder and deep decoder'' architecture instead of encoder and decoder of the same size in conventional T5 models~\citep{t5, codet5}.
As noted by \citet{alphacode}, the decoder in a T5-based model is often required to deal with a higher level of complexity in generation tasks and thus, should be enhanced with a larger number of neural parameters.

To connect the separately pretrained encoder and decoder, we insert  randomly initialized cross-attention layers to decoder blocks after the self-attention layers.
For the purpose of efficient tuning, we  only insert  cross-attention layers to  the top-$L$  decoder layers ($L$=1 in our experiments).
We  only keep the small encoder and cross-attention layers trainable while freezing the majority of the decoder parameters.   
We also explored other advanced designs such as adding a gating function to improve training stability or inserting multiple cross-attention layers at a certain frequency ~\citep{flamingo}.
However,  we did not observe significant performance improvement, and worse still, these design choices would introduce too expensive computation overhead.

\subsection{Adaptation to Downstream Understanding and Generation Tasks}
\label{subsec:finetuning}
After the two stages of pretraining, CodeT5+ can  flexibly operate in various modes to support different tasks, including Seq2Seq  generation tasks, decoder-only tasks, and  understanding-based tasks:

\textbf{Seq2Seq Generation Tasks.} 
As an encoder-decoder model, CodeT5+ can be naturally adapted to a variety of Seq2Seq generation tasks such as code generation  and  summarization.
We also adapt CodeT5+ as a retrieval-augmented generation model, using the encoder to retrieve code snippets, which are then used by both the encoder and decoder for code generation.

\textbf{Decoder-only Tasks.} 
In this setting, we always feed a \texttt{[CLM]} token to the encoder input and pass the source sequence to the decoder as the prefix context.
We freeze the weights of the encoder and the cross-attention layers in the decoder.
This strategy only activates parts of the decoder and technically reduces about half of the total model parameters. 
We use next-line code completion tasks to evaluate the decoder-only generation capability of CodeT5+.

\textbf{Understanding Tasks.} 
CodeT5+ can support these understanding tasks in two ways: first, it employs the encoder to obtain text/code embeddings, which can be either passed to a binary classifier for detection tasks or retrieval tasks;
alternatively, the encoder can be combined with the decoder to predict the text-code matching scores for text-to-code retrieval tasks.

\section{Pretraining and Instruction Tuning}\label{app_sec:pretrain}

\subsection{Pretraining Dataset}
We enlarge the pretraining dataset  of CodeSearchNet~\citep{csn} with the recently released GitHub Code dataset\footnote{\url{https://huggingface.co/datasets/codeparrot/github-code}}. We select nine PLs (Python, Java, Ruby, JavaScript, Go, PHP, C, C++, C\#) and filter the dataset by preserving only permissively licensed code\footnote{Permissive licenses: “mit”, “apache-2”, “bsd-3-clause”, “bsd-2-clause”, “cc0-1.0”, “unlicense”, “isc”} and files with  50 to 2000 tokens. Besides, we filter out the overlapped subset with CodeSearchNet and other downstream tasks covered in our evaluation by checking their GitHub repository names. 
Note that although we employ the deduplicated data version in which duplicates are filtered out based on the exact match (ignoring whitespaces), there might be some potential remaining duplicates. However, we do not expect any remaining duplication will impact our model performance significantly.
We use the CodeT5 tokenizer to tokenize the multilingual dataset, resulting in 51.5B tokens, $\sim$50x larger than CodeSearchNet.

\begin{table}[t]
 \centering
 
\caption{Data statistics of both unimodal and bimodal (CodeSearchNet) pretraining data}
\label{tab:data_stat_pretrain}

\begin{tabular}{llrp{3.2cm}}
\hline
Dataset                        & Language   &   \# Sample         &   Total size                                   \\
\hline

\multirow{9}{*}{Ours}           & Ruby       & 2,119,741  & \multirow{9}{3.2cm}{37,274,876 files}     \\

                               & JavaScript & 5,856,984  &                                                             \\
                               & Go         & 1,501,673  &                                                              \\
                               & Python     & 3,418,376  &                                                             \\
                               & Java       & 10,851,759 &                                                              \\
                               & PHP        & 4,386,876  &                                                              \\
                               & C          & 4,187,467  &                                                              \\
                               & C++        & 2,951,945  &                                                              \\
                               & C\#        & 4,119,796  &                                           \\
       \hline                        
\multirow{6}{*}{CSN} & Ruby       & 49,009    & \multirow{6}{3.2cm}{1,929,817 text-code pairs at function level}   \\

                               & JavaScript & 125,166  &                                                              \\
                               & Go         & 319,132    &                                                              \\
                               & Python     & 453,772  &                                                              \\
                               & Java       & 457,381  &                                                              \\
                               & PHP        & 525,357    &                                                              \\

\hline                               
\end{tabular}

\end{table}
We report the data statistics of both unimodal code  and bimodal text-code  pretraining datasets in \cref{tab:data_stat_pretrain}.
From the table, we can see that our curated dataset from GitHub code has a much larger data size at the file level than the CodeSearchNet bimodal data at the function level, allowing our model to learn rich representations in the first stage of pretraining.
Different from CodeT5~\citep{codet5} which employs both unimodal and bimodal data in CodeSearchNet~\cite{csn}, we only employ its bimodal subset for the second stage pretraining of our CodeT5+.
 We use this stage to mainly adapt our model to text-code related tasks like text-to-code retrieval and generation.

\subsection{Pretraining Setup}\label{subsec:setup}
We pretrained two groups of CodeT5+ models: 1) CodeT5+ 220M and 770M that are trained from scratch following T5's architecture~\citep{t5} (T5-base and large respectively), 2) CodeT5+ 2B, 6B, 16B in which the decoders are initialized from CodeGen-mono 2B, 6B, 16B models~\citep{codegen} and its encoders are initialized from CodeGen-mono 350M. Note that following our model scaling strategy, the latter group of CodeT5+ models introduce insignificant trainable parameters (the 350M encoder plus one cross-attention layer of 36M, 67M, 151M for 2B, 6B, 16B models respectively) compared to the original CodeGen models.
%, but their overall model sizes still at a  similar size.
We employ the CodeT5 tokenizer and CodeGen tokenizer for these two groups of models respectively.
In pretraining, we adopt a stage-wise strategy to  pretrain CodeT5+ first on the large-scale unimodal dataset and then on the smaller bimodal dataset  on  a cluster  with 16 A100-40G GPUs on Google Cloud Platform.

In the first stage, we  warm up the model with the span denoising task for $10k$ training steps, and then joint training with the two CLM tasks with equal weights for $100k$ steps. We employ a linear decay learning rate (LR) scheduler with a peak learning rate of 2e-4 and set the batch size to 2048 for denoising and 512 for CLM.
To prepare the input and output data, we set the maximum length to $512$ for the denoising task,  and set the maximum lengths to $768$ and $600$ for source and target sequences for the code completion CLM, $1$ and $1024$ for the decoder-only generation CLM.
% To further speed up training, we concatenate data samples and truncate them into blocks of fixed size of 512 for the span denoising tasks, while we feed single files for CLM tasks.
In the second stage, we jointly optimize four losses of contrastive learning, matching, and two CLM losses with equal weights for $10$ epochs with a batch size of $256$.  We  employ a peak learning rate of 1e-4 and set the maximum sequence lengths to $420$ and $128$ for code and text sequences.

In all  experiments, we employ an AdamW optimizer~\citep{DBLP:conf/iclr/LoshchilovH19} with a $0.1$ weight decay. We also employ the DeepSpeed's ZeRO Stage 2~\citep{deepspeed} with mixed precision training of FP16 for training acceleration. For the training of CodeT5+ 2B, 6B, and 16B, we use FP16 frozen decoder weights and keep other trainable weights in FP32. We use DeepSpeed ZeRO Stage 3's parameter partition for CodeT5+ 6B and 16B models.

\subsection{Instruction Tuning}\label{subsec:instruct}

\begin{figure*}[t]
	\centering
	\resizebox{1.0\textwidth}{!} {
 \includegraphics{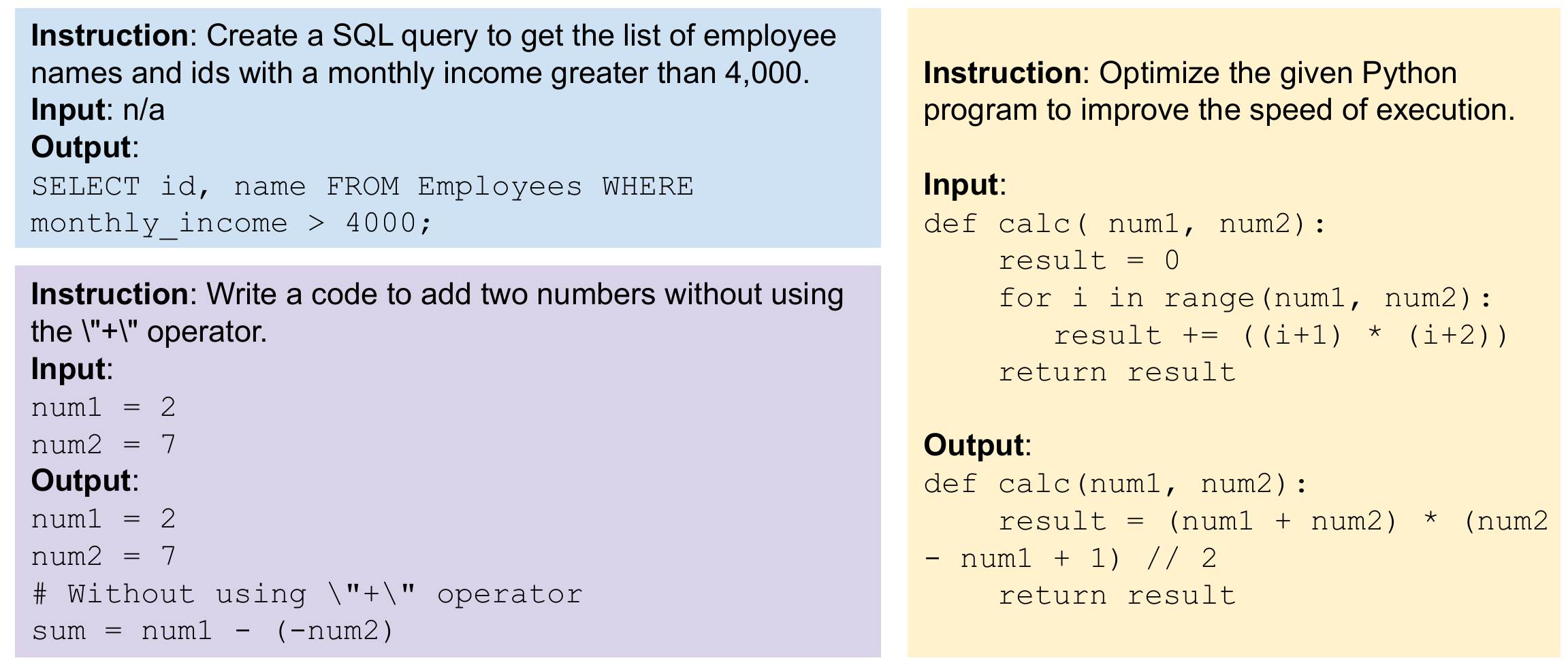}
	}
	\caption{
        \textbf{Example generated instruction data:}
        we demonstrate some examples of instruction data used to finetune CodeT5+ to better align our models to natural language instructions. 
        The instruction corpus contains novel tasks, such as text-to-SQL generation and Python code optimization. 
        }
	\label{fig:sample_instructions}
\end{figure*}

In the NLP domain, recent work \citep{wang2022self, alpaca} studied the benefits of data augmentation techniques on pretrained LMs with synthetic instruction data. 
Models finetuned with this type of data can better understand natural language instructions and demonstrate improved alignment with the corresponding tasks \citep{wang2022self, ouyang2022training}. 
We are motivated to transfer this technique to the code domain to improve our CodeT5+ models.
Following \citet{alpaca}, we employ over $20$k instruction data in the code domain curated by~\citet{codealpaca}.
The data is generated by letting pretrained LLMs i.e. text-davinci-003, generate novel tasks, including task instructions, inputs (if any), and expected outputs. 
We trained our models on this augmented dataset for up to 3 epochs and denote the instruction-tuned models as ``InstructCodeT5+''. 
Note that the instruction data are generated fully independently from any downstream evaluation tasks and we still evaluate the instruction-tuned models in a zero-shot manner. 
Fig. \ref{fig:sample_instructions} illustrates some examples of the generated instruction data.
Note that as we rely on LM-generated data, including the annotations of expected outputs, not all of the data is perfectly correct. 
For instance, the example of the code optimization task in Fig. \ref{fig:sample_instructions} contains a wrong output. 
\citet{wang2022self} treated these examples as data noise and the tuned models still benefit from the majority of the synthetic instruction dataset. 

\section{Experiments}
We conducted comprehensive experiments on a wide range of code understanding and generation tasks over 20+ code-related datasets across 9 different programming languages (PLs). In addition, we consider a variety of evaluation settings including zero-shot, instruction tuning, task-specific finetuning.
Additional results and detailed finetuning setups can be found in the Appendix~\ref{app_sec:more_result} and \ref{app_sec:finetune}.

\textbf{Baselines.} 
We implemented a family of  CodeT5+ models, with model sizes ranging from 220M to 16B. 
Note that CodeT5+ 220M and 770M employ the same architecture of T5 \citep{t5} and are pretrained from scratch, while CodeT5+ 2B, 6B, 16B employ the ``shallow encoder and deep decoder'' architecture with encoders initialized from CodeGen-mono 350M and decoders initialized from CodeGen-mono 2B, 6B, 16B, respectively.
We compare CodeT5+ with state-of-the-art code LLMs that can be  categorized into $3$ types: encoder-only, decoder-only, and encoder-decoder models.
\begin{itemize}[leftmargin=3mm]
    \item For \textbf{encoder-only} models, we consider RoBERTa~\citep{roberta},  CodeBERT~\citep{codebert} trained with  masked language modeling,
GraphCodeBERT~\citep{graphcodebert} using data flow extracted from abstract syntax tree (AST) of code,  SYNCOBERT~\citep{syncobert} and UniXcoder~\citep{unixcoder} that incorporates AST and contrastive learning. Note that  UniXcoder can be also viewed as decoder-only model as it employs UniLM-style masking~\citep{unilm}.  

    \item For \textbf{decoder-only} models, we consider GPT-2~\citep{gpt} and CodeGPT~\citep{codexglue}. Both are pretrained using a CLM objective. 
    In this model paradigm, we also consider models of very large scales (up to 540B parameters) such as  PaLM \citep{palm},  GPT-4~\citep{OpenAI2023GPT4TR}, Codex \citep{codex}, LLaMA \citep{touvron2023llama}, CodeGen~\citep{codegen}, Incoder~\citep{incoder}, GPT-J \citep{gpt-j}, GPT-Neo and GPT-NeoX \citep{gpt-neox-20b}, MIM \citep{nguyen2023meet}, CodeGeeX \citep{zheng2023codegeex}. We also compare with Replit~\citep{replit} and StarCoder~\citep{starcoder} which are released concurrently with this work. 

    \item For \textbf{encoder-decoder} models, we consider PLBART~\citep{plbart} and CodeT5~\citep{codet5}, which employ a unified framework to support understanding and generation tasks.

\end{itemize}

Note that billion-parameter LLMs such as Codex and CodeGen typically use most of the source code from GitHub for model training  and do not remove any overlap with the downstream tasks covered in this work as we did.
Therefore, it is difficult to ensure a fair comparison with these models in those tasks, especially the  code summarization and completion tasks. 
Moreover, these models are very expensive to perform task-specific finetuning, and hence, they are often employed only on the zero-shot evaluation.
In this work, we mainly compare CodeT5+ with  these LLMs in the zero-shot HumanEval code generation task~(\cref{subsec:humaneval}).
In other experiments, we focus on the finetuning setting and compare our models with smaller-scale LMs, including CodeGen-multi-350M despite its potential data leakage issues during pretraining.
In some of the finetuning evaluations such as  the code summarization (\cref{subsec:summarize}) and text-to-code retrieval tasks (\cref{subsec:retrieve}), we found that the performance improvement already becomes relatively saturated as the model size increases. 
%of CodeT5+ increases from  220M to 770M.
This implies that with enough data for finetuning, these tasks might not benefit much from model scaling (to billions of parameters) as compared to the zero-shot evaluation settings.

\begin{table}[htbp]
\centering

\caption{
\textbf{Results of \emph{pass@k}(\%) on HumanEval:}
We compare our models with 
\emph{(i)} closed-source models (top) such as AlphaCode \citep{alphacode}, Codex \citep{codex}, and GPT-4 \citep{OpenAI2023GPT4TR}; 
\emph{(ii)} open-source models (middle) such as CodeGen \citep{codegen}, Incoder \citep{incoder}, and LLaMA \citep{touvron2023llama}; and 
\emph{(iii)} models with enhancement generation strategies (bottom) such as unit test generation \citep{chen2023codet} and prompting.
}
\label{tab:humaneval}

\begin{tabular}{lcccc}
\hline
Model                        & Model size & pass@1 & pass@10 & pass@100 \\
\hline

\multicolumn{5}{c}{{Closed-source models}}       \\ \hline
LaMDA                        & 137B & 14.0  & -       & 47.3    \\
AlphaCode     & 1.1B & 17.1  & 28.2   & 45.3    \\
MIM                          & 1.3B & 22.4  & 41.7   & 53.8    \\
MIM                          & 2.7B & 30.7  & 48.2   & 69.6    \\
%\hdashline
PaLM                         & 8B   & 3.6   & -       & 18.7    \\
PaLM                         & 62B  & 15.9  & -       & 46.3    \\
PaLM                         & 540B & 26.2  & -       & 76.2    \\
PaLM-Coder                   & 540B & 36.0  & -       & 88.4    \\
%\hdashline
Codex                        & 2.5B & 21.4  & 35.4   & 59.5    \\
Codex                        & 12B  & 28.8  & 46.8   & 72.3    \\

code-cushman-001             & -    &33.5   &54.3   &77.4 \\
code-davinci-002             & -    & 47.0  & \textbf{74.9}   & \textbf{92.1}    \\
GPT-3.5                      & -    & 48.1  & -       & -        \\
GPT-4                        & -    & \textbf{67.0}  & -       & -        \\
\hline

\multicolumn{5}{c}{{Open-source models}}  \\ \hline
% GPT-Neo                      & 1.3B & 4.8   & 7.5    & 16.3    \\
GPT-Neo                      & 2.7B & 6.4   & 11.3   & 21.4    \\
GPT-J                        & 6B   & 11.6  & 15.7   & 27.7    \\
GPT-NeoX                     & 20B  & 15.4  & 25.6   & 41.2    \\

InCoder                      & 1.3B & 8.9   & 16.7   & 25.6    \\
InCoder                      & 6B   & 15.2  & 27.8   & 47.0    \\

CodeGeeX                     & 13B  & 22.9  & 39.6   & 60.9    \\

LLaMA                        & 7B   & 10.5  & -       & 36.5    \\
LLaMA                        & 13B  & 15.8  & -       & 52.5    \\
LLaMA                        & 33B  & 21.7  & -       & 70.7    \\
LLaMA                        & 65B  & 23.7  & -       & \textbf{79.3}    \\

Replit                        & 3B  & 21.9  & -       & -    \\

StarCoder                        & 15B  & 33.6  & -       & -    \\

% CodeGen-mono                 & 350M & 12.8  & 23.1   & 35.2    \\
CodeGen-mono                 & 2B   & 23.7  & 36.6   & 57.0    \\
CodeGen-mono                 & 6B   & 26.1  & 42.3   & 65.8    \\
CodeGen-mono                 & 16B  & 29.3  & 49.9   & 75.0    \\
\hdashline

CodeT5+                       & 220M & 12.0  & 20.7   & 31.6    \\
CodeT5+                       & 770M & 15.5  & 27.2   & 42.7    \\
CodeT5+                       & 2B   & 24.2  & 38.2   & 57.8    \\
CodeT5+                       & 6B   & 28.0  & 47.2   & 69.8    \\
CodeT5+                       & 16B  & {30.9}  & {51.6}   & {76.7}    \\
InstructCodeT5+            & 16B  & \textbf{35.0}  & \textbf{54.5}   & 77.9    \\

\hline

\multicolumn{5}{c}{{Open-source models + generation strategies}}             \\ 
\hline

StarCoder (prompted)                      & 15B  & 40.8  & -       & -    \\
CodeGen-mono w/ CodeT         & 16B  & 36.7  & 59.3   & -        \\
\hdashline
CodeT5+ w/  CodeT & 16B  & 38.5   & 63.6   &   77.1 \\
InstructCodeT5+ w/ CodeT & 16B  & \textbf{42.9}  & \textbf{67.8}  &  \textbf{78.7} \\
\hline
\end{tabular}
\end{table}

\subsection{Zero-shot Evaluation on Text-to-Code Generation Tasks}~\label{subsec:humaneval}

We first evaluate the model capabilities to generate Python code given natural language specifications in a zero-shot setting. 
In this task, from CodeT5+, we activate both encoder and decoder modules, whereby the encoder encodes an input text sequence and the decoder generates corresponding programs conditioned on the input text. 
We use the HumanEval benchmark \citep{codex}, which consists of 164 Python problems.
To evaluate models for code generation, exact match or BLEU scores might be limited as there can be multiple versions of correct program solutions. 
Besides, \citet{codex} found that the functional correctness of generated codes correlates poorly with their BLEU scores. 
Therefore, we evaluate the model performance by testing generated codes against unit tests. 
We report the passing rate \emph{pass@k} in Table \ref{tab:humaneval}.
Following prior approaches in this benchmark, we adopted nucleus sampling during inference with a temperature of $0.2$, $0.6$, and $0.8$ for $k=\{1,10,100\}$. 
In this experiment, we follow \citet{codegen} to continue to pretrain our CodeT5+ models on another epoch of the Python  subset data using causal LM objective to better adapt them for Python code generation.

In the zero-shot setting, the  instruction-tuned CodeT5+ ("InstructCodeT5+") 16B model can improve the performance against other open code LLMs, achieving new SoTA of $35.0\%$  pass@1 and  $54.5\%$  pass@10.
Its pass@100 result of $77.9\%$ is not too far behind the current SoTA open-source model, i.e. LLaMA 65B \citep{touvron2023llama}.
Particularly, as an open model, our InstructCodeT5+ 16B even outperforms the OpenAI code-cushman-001 model across all metrics.
We also observed that our small-sized models of 220M and 770M already match or outperform much larger code LLMs, e.g.,   CodeT5+ 770M's $15.5\%$ pass@1 compared to Incoder 6B's $15.2\%$, GPT-NeoX 20B's $15.4\%$, and PaLM 62B's $15.9\%$.  
Besides,  we observed that compared to the CodeGen models of similar sizes \citep{codegen}, CodeT5+ obtains consistent performance gains from 2B to 16B model variants.  
These superior results against decoder-only baselines demonstrate the advantage of the encoder-decoder architecture of CodeT5+ and validate the effectiveness of our proposed compute-efficient pretraining strategy with frozen off-the-shelf code LLMs.  

Finally, we evaluated the models with enhancement generation strategies following CodeT \citet{chen2023codet}.
In this setting, we let models generate additional test cases (by prompting the models with an \texttt{assert} statement). 
We then used these generated test cases to filter and sample generated code samples for evaluation. 
We observe that this strategy can select better code candidates and bring the performance gains, achieving up to $42.9\%$ pass@1 and  $67.8\%$ pass@10.
We do notice the performance gaps of CodeT5+ against closed-source models such as GPT-4 \citep{OpenAI2023GPT4TR} and code-davinci-002. 
However, as the implementation details and model weights/sizes of these models were not released, it is difficult to diagnose the root causes of the performance gaps.

\subsection{Evaluation on Math Programming Tasks}

\begin{figure}[t]
	\centering
	\resizebox{\textwidth}{!} {
	\includegraphics{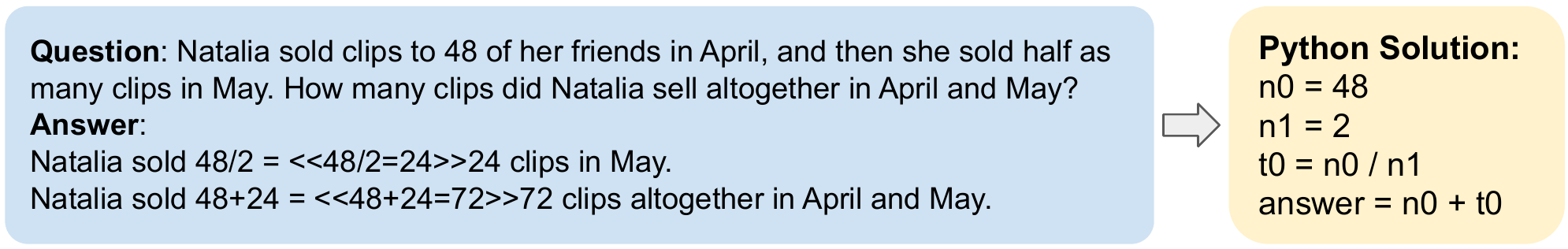}
	}
	\caption{
\textbf{GSM8K benchmark:}    
 One example of how to convert natural language solution into a Python program on GSM8K dataset.}
	\label{fig:gsm_one_case}
\end{figure}

We consider other code generation tasks, specifically 
two math programming benchmarks MathQA-Python~\citep{austin2021program} and GSM8K~\citep{gsm}.
The task is to generate Python programs to solve mathematical problems described in natural language descriptions, where  code correctness is measured based on the execution outputs of the generated programs. 
We follow~\citep{austin2021program} to convert the solutions in GSM8K into  Python programs (henceforth GSM8K-Python, one example is illustrated in \cref{fig:gsm_one_case}). 
We employ {pass@k}, to measure the percentage of problems solved using $k$ generated programs per problem.
We compare our models with very large-scale decoder-only models including LaMDA~\citep{austin2021program}, LLaMA~\citep{touvron2023llama}, Minerva \citep{lewkowycz2022solving}, code-davinci~\citep{codex},  GPT-Neo~\citep{gptneo}, and CodeGen~\citep{codegen}.
Some of the prior approaches  are enhanced with generation strategies such as self-sampling optimization~\citep{sampleprogram} and majority voting \citep{lewkowycz2022solving}.

\begin{table}[t]
\centering
\caption{
\textbf{Results of \emph{pass@k}(\%) on MathQA-Python and GSM8K-Python:}
Despite the smaller model size, CodeT5+ models can outperform other much larger language models.
Note that some baseline models adopt additional generation strategies, such as GPT-Neo using self-sampling optimization \citep{sampleprogram}, and LLaMA and Minerva using majority voting \cite{lewkowycz2022solving}
}
\label{tab:math}
\begin{tabular}{lccc}
\hline
\multicolumn{1}{c}{Model} & Model size & MathQA-Python & GSM8K-Python \\
                          &            & pass@80       & pass@100     \\
\hline
\multicolumn{4}{c}{Few-shot learning results}\\ \hline 
code-davinci              & -          & \textbf{42.0}          & 71.0         \\ 
%\hdashline
LLaMA                     & 13B        & -             & 29.3         \\
LLaMA                     & 33B        & -             & 53.1         \\
LLaMA                     & 65B        & -             & 69.7         \\
%\hdashline
Minerva                   & 8B         & -             & 28.4         \\
Minerva                  & 62B        & -             & 68.5         \\
Minerva                   & 540B       & -             & \textbf{78.5}         \\ 
\hline 
\multicolumn{4}{c}{Finetuning results}\\ \hline
LaMDA                     & 137B       & 81.2          & -            \\ 
%\hdashline
GPT-Neo                   & 125M       & 84.7          & -            \\
GPT-Neo                  & 2.7B       & -             & 41.4         \\ 
%\hdashline
CodeGen-mono              & 350M       & 83.1          & 38.7         \\
CodeGen-mono              & 2B       & 85.6          & 47.8         \\

CodeT5                    & 220M       & 71.5          & 58.4         \\
\hdashline
CodeT5+                   & 220M       & 85.6          & 70.5         \\
CodeT5+                   & 770M       & \textbf{87.4}          & \textbf{73.8}        \\
\hline 
\end{tabular}
\end{table}
\begin{figure}[t]
	\centering
	\resizebox{0.8\textwidth}{!} {
	\includegraphics{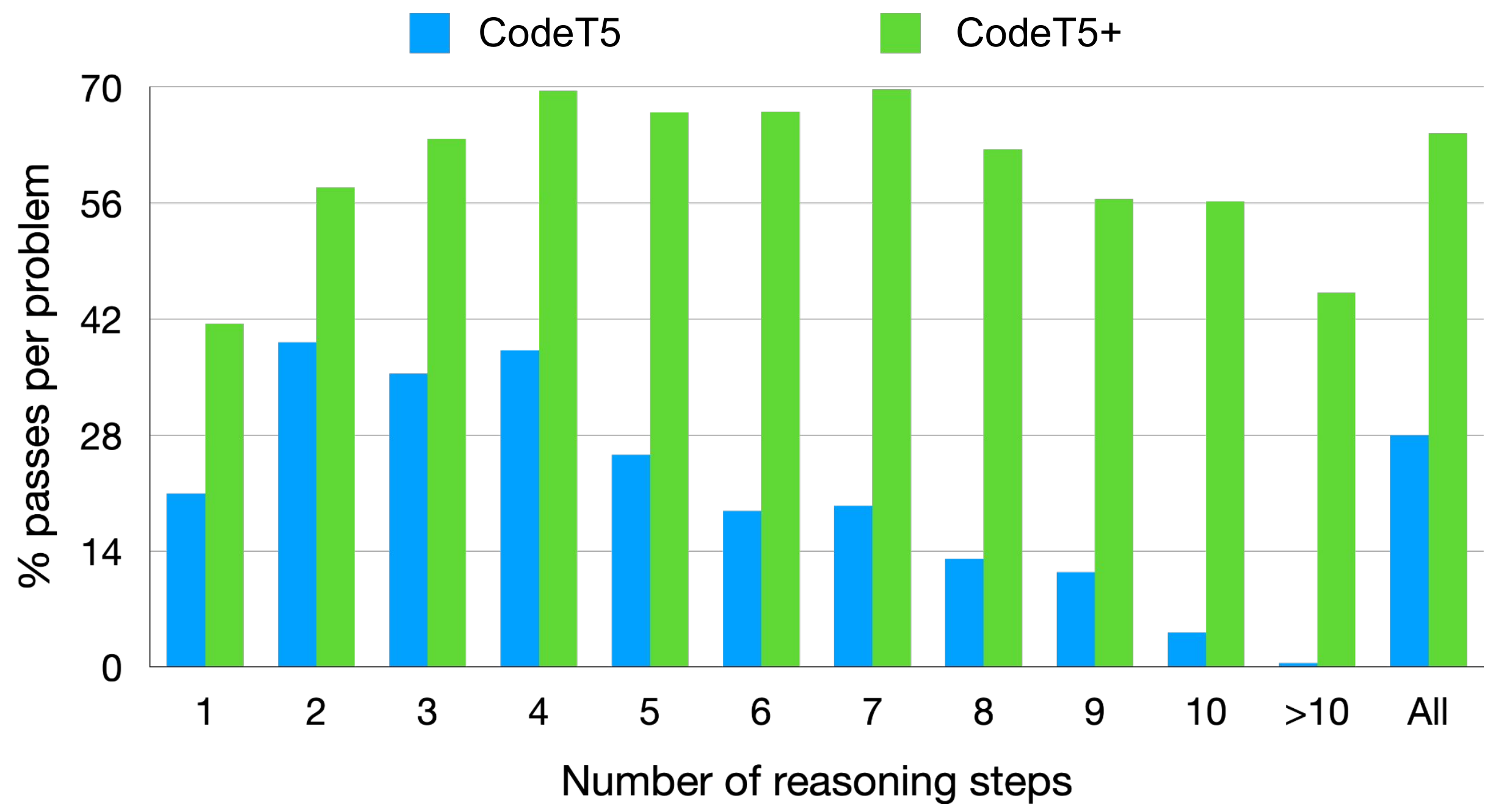}
	}
	\caption{
        \textbf{Results of MathQA-Python programming tasks by problem complexity:}
        compared to CodeT5, CodeT5+ is more robust against the complexity of the problems (i.e. the number of reasoning steps required). 
        This observation demonstrates improved reasoning capabilities of CodeT5+, in addition to its understanding and generation skills.
    }
	\label{fig:math_programming_qualitative}
\end{figure}

\cref{tab:math} shows that CodeT5+ achieves significant performance gains, outperforming many code LLMs of much larger sizes. 
Specifically, our CodeT5+ 770M achieves new SoTA results of $87.4$ {pass@80} on MathQA-Python and very competitive results of $73.8$ {pass@100} on GSM8K-Python. 
On GSM8K-Python, CodeT5+ 770M achieves the best finetuning results against other larger models (e.g., GPT-Neo 2.7B and CodeGen-mono 2B),  and outperforms LaMDA 137B and code-davinci in the few-shot evaluation setting.
We did observe that our models still have some performance gap against Minerva \citep{lewkowycz2022solving}.
Note that this model was initialized with pretrained PaLM  \citep{palm} and further finetuned with large-scale scientific corpora. 
The model also employs a majority voting strategy to select the most common answers as the final predictions. 

In \cref{fig:math_programming_qualitative}, we analyze the model performance by the complexity of math programming problems on MathQA-Python. 
For each problem, we extract the number of reasoning steps required to solve the problem.
We observe that compared to CodeT5, CodeT5+ is more robust against the complexity of the problems (i.e. the number of reasoning steps required). 
CodeT5 model performance tends to deteriorate drastically as the number of reasoning steps increases. 
In CodeT5+, the downward trend is a lot less severe and the model still achieves good results in very complex tasks (more than $10$ steps).
Please see Appendix~\ref{app_ssec:math_qualitative} for more qualitative examples.

\subsection{Evaluation on Code Summarization Tasks}\label{subsec:summarize}

\begin{table*}[t]
\centering

\caption{
\textbf{Results (smoothed BLEU-4)  on code summarization on CodeSearchNet:}
CodeT5+ can generate better code summaries across all 6 programming languages, outperforming strong baselines such as UniXcoder and CodeT5. 
}
\label{tab:code_sum_full}
\begin{tabular}{l|c c c c c c|c}
\hline
Model & Ruby & JS & Go & Python & Java & PHP & Overall \\
\hline
RoBERTa 125M        & 11.17 & 11.90      & 17.72 & 18.14  & 16.47 & 24.02 & 16.57   \\
CodeBERT 125M       & 12.16 & 14.90      & 18.07 & 19.06  & 17.65 & 25.16 & 17.83   \\

UniXcoder 125M       & 14.87 & 15.85      & 19.07 & 19.13  & 20.31 & 26.54 & 19.30   \\
CodeGen-multi 350M &13.48&16.54&18.09&18.31&19.41&24.41&18.37 \\
\hline
PLBART 140M         & 14.11 & 15.56      & 18.91 & 19.30  & 18.45 & 23.58 & 18.32   \\
CodeT5 220M     & 15.24 & 16.16      & 19.56 & 20.01  & 20.31 & 26.03 & 19.55   \\

\hline
CodeT5+ 220M  & 15.51 & 16.27      & 19.60  & 20.16  & 20.53 & 26.78 & 19.81   \\
CodeT5+ 770M & 15.63 & 17.93      & 19.64 & 20.47  & 20.83 & 26.39 & \textbf{20.15}  \\
\hline
\end{tabular}
\end{table*}

The code summarization task aims to summarize a code snippet into natural language docstrings. 
We employ the clean version of CodeSearchNet dataset \citep{csn} in six programming languages to evaluate our models for this task.
We employ BLEU-4 \citep{bleu} as the performance metric which measures the token-based similarity between predicted and ground-truth summaries.  
From pretrained CodeT5+, we activate both encoder and decoder for this task.

From \cref{tab:code_sum_full}, we found that encoder-decoder models (CodeT5 and CodeT5+) generally outperform both encoder-only models \citep{codebert} and decoder-only models \citep{codegen}, as well as the UniLM-style model UniXcoder \citep{unixcoder}.
This observation demonstrates the benefit of using the encoder-decoder architecture in CodeT5+ to better encode code contexts and generate more accurate code summaries. 
Finally, we also observed some performance gains against CodeT5 \citep{codet5}, indicating the advantage of our proposed mixture of diverse pretraining learning objectives in addition to the span denoising objective in CodeT5.

\subsection{Evaluation on Code Completion Tasks}

\begin{table}[t]
\centering
\caption{
\textbf{Results on line-level code completion on PY150 and JavaCorpus:}
The flexibility of CodeT5+ is demonstrated in its adaptation as a decoder-only model for code completion tasks. 
Our models is able to outperform many strong GPT-based baselines such as CodeGen-multi and UniXcoder.
EM: Exact Match, Edit Sim:  Levenshtein edit similarity. 
}
\label{tab:code_comp}

\begin{tabular}{l|cc|cc}
\hline
\multirow{2}{*}{Model} & \multicolumn{2}{c|}{PY150} & \multicolumn{2}{c}{JavaCorpus} \\
                       & EM         & Edit Sim     & EM           & Edit Sim        \\
\hline
CodeGPT 124M               & 42.37      & 71.59        & 30.60        & 63.45           \\
UniXcoder 125M             & 43.12      & 72.00        & 32.90        & 65.78           \\
CodeGen-multi 350M & 42.47 & 70.67 & 35.47 & 69.22 \\
\hline
PLBART 140M                & 38.01      & 68.46        & 26.97        & 61.59           \\
CodeT5 220M            & 36.97      & 67.12        & 24.80        & 58.31           \\

\hline
CodeT5+ 220M         & 43.42      & 73.69        & 35.17        & 69.48           \\
CodeT5+ 770M        & \textbf{44.86}      & \textbf{74.22}        & \textbf{37.90}        & \textbf{72.25}      \\

\hline
\end{tabular}

\end{table}

We evaluate the  decoder-only generation capability of CodeT5+ through a line-level code completion task, which aims to complete the next  code line based on the previous code contexts. 
We employ PY150~\citep{py150} and GitHub JavaCorpus~\citep{javacorpus} from CodeXGLUE, and use  exact match (EM) accuracy and Levenshtein edit similarity~\citep{editsim} as evaluation metrics. 
In this task, we employ a {decoder-only} model from CodeT5+ so that only about half of the total model parameters are activated.  

\cref{tab:code_comp} shows that both CodeT5+ (in decoder-only mode) and  decoder-only models (the top block) significantly outperform encoder-decoder models (the middle block), validating that decoder-only models can better suit the  code completion task in nature.
Specifically, CodeT5+ 220M already surpasses UniXcoder and achieves comparable performance to CodeGen-multi 350M, while the 770M one further sets new SoTA results in both metrics.
In particular, CodeT5+ 220M yields substantial improvements over CodeT5 model of the same size by +6.5 EM and +10.4 EM scores on PY150 and JavaCorpus respectively. This is mainly due to our  causal LM objectives in the first-stage pretraining, which allows the decoder to see longer sequences instead of a combination of discrete spans in CodeT5, leading to a better causal  generation capability.

\subsection{Evaluation on Text-to-Code Retrieval Tasks}\label{subsec:retrieve}

\begin{table} [t]
\caption{
\textbf{Text-to-Code Retrieval results (MRR) on CodeXGLUE:}
CodeT5+ achieves consistent performance gains over the original CodeT5 models across all $3$ retrieval benchmarks in $7$ programming languages. 
Overall, our models demonstrate remarkable performance, outperforming many strong encoder-based  models pretrained with contrastive loss such as SYNCOBERT and UniXcoder.  
}
\label{tab:code_search}

\resizebox{1.0\linewidth}{!}{
\begin{tabular}{l|c c c c c c|c|c|c}
\hline
\multirow{2}{*}{Model} & \multicolumn{7}{c|}{CodeSearchNet}                         & \multirow{2}{*}{CosQA} & \multirow{2}{*}{AdvTest} \\
\cline{2-8}
                       & Ruby & JS & Go   & Python & Java & PHP  & Overall &                        &                          \\
\hline
% RoBERTa 125M               & 58.7 & 51.7       & 85.0 & 58.7   & 59.9 & 56.0 & 61.7    & 60.3                   & 18.3                     \\
CodeBERT 125M              & 67.9 & 62.0       & 88.2 & 67.2   & 67.6 & 62.8 & 69.3    & 65.7                   & 27.2                     \\
GraphCodeBERT 125M          & 70.3 & 64.4       & 89.7 & 69.2   & 69.1 & 64.9 & 71.3    & 68.4                   & 35.2                     \\
SYNCOBERT  125M           & 72.2 & 67.7       & 91.3 & 72.4   & 72.3 & 67.8 & 74.0    & -                      & 38.3                     \\
UniXcoder  125M             & 74.0 & 68.4       & 91.5 & 72.0   & 72.6 & 67.6 & 74.4    & 70.1                   & 41.3                     \\
CodeGen-multi 350M & 66.0 &	62.2 &	90.0 &	68.6	&70.1	& 63.9	&70.1 & 64.8 & 34.8 \\
\hline
PLBART 140M                 & 67.5 & 61.6       & 88.7 & 66.3   & 66.3 & 61.1 & 68.6    & 65.0                   & 34.7                     \\

CodeT5 220M            & 71.9 & 65.5       & 88.8 & 69.8   & 68.6 & 64.5 & 71.5    & 67.8                   & 39.3                     \\

\hline
CodeT5+ 220M        & 77.7 & 70.8 & 92.4 & 75.6 & 76.1 & 69.8 & 77.1    & 72.7                   & 43.3                     \\
CodeT5+ 770M      & \textbf{78.0} & \textbf{71.3} & \textbf{92.7} & \textbf{75.8} & \textbf{76.2} & \textbf{70.1} & \textbf{77.4}    & \textbf{74.0}                      & \textbf{44.7}                       \\
\hline
\end{tabular}
}
\end{table}

We evaluate the code understanding capabilities of CodeT5+ through text-to-code retrieval tasks across multiple PLs.
This task aims to find the most semantically related code snippet at the function level from a collection of candidate codes based on a natural language query. We consider three datasets for evaluation: CodeSearchNet~\citep{csn}, CosQA~\citep{cosqa}, and AdvTest~\citep{codexglue}, which are curated from the original CodeSearchNet by  filtering data with low-quality queries, adopting real-world queries from a modern search engine, and  obfuscating identifiers to normalize the code.
In this task, we activate both encoder and decoder  of CodeT5+
and use Mean Reciprocal Rank (MRR) as the evaluation metric.

From~\cref{tab:code_search}, our CodeT5+ 220M significantly outperforms all existing  encoder-only/decoder-only (the top block) and encoder-decoder models (the middle block).
Our CodeT5+ 770M further sets new SoTA results, surpassing the previous SoTA UniXcoder by more than 3 absolute MRR points on all 3 tasks across 8 datasets. This implies CodeT5+ is a robust  code retriever model to  handle queries with diverse formats and PLs. 
Besides, CodeT5+ 220M yields substantial performance gains over  CodeT5 model of the same size.
These gains can be attributed to the text-code contrastive learning and matching objectives that facilitate better unimodal and bimodal representation learning.
Particularly, compared to  SYNCOBERT and UniXcoder also pretrained  with contrastive learning, our models achieve much better results, which can be attributed to our text-code matching pretraining task that enables the exploitation of  more fine-grained text-code alignments.

\subsection{Ablation Study}
\label{subsec:ablation}

\begin{table}[t]
%\vspace{-0.05in}
%\small
\centering
\caption{
\textbf{Ablation results of CodeT5+:}
a) no causal LM objective during stage-1 unimodal pretraining,
b) no matching or causal LM objective during stage-2 bimodal pretraining. 
}

\label{tab:ablation}
 \begin{tabular}{l|cc|cc}
 \hline
\multirow{3}{*}{Model} & \multicolumn{2}{c|}{Code Completion} & \multicolumn{2}{c}{Math Programming} \\
\cline{2-5}  
                       & PY150          & JavaCorpus         & MathQA-PY         & GSM8K-PY         \\
%\cline{2-5}                       
                       & EM             & EM                 & pass@80           & pass@100         \\
\hline
CodeT5+ 770M        & \textbf{44.9}          & \textbf{37.9}              & \textbf{87.4}              & \textbf{73.8}             \\
\hline
a)~~~no causal LM                 & 36.2             & 24.8                 & 72.3             & 61.4    \\
 \hline
 \end{tabular}

\hfill

\vspace{0.1in}

 \begin{tabular}{p{0.1cm}l|cccccc|c}
 \hline
  & \multirow{2}{*}{Model} & \multicolumn{7}{c}{Text-to-code Retrieval} \\
   \cline{3-9}
 & & Ruby & JS & Go   & Python & Java & PHP  & Overall  \\
\hline
 & CodeT5+ 770M       & \textbf{78.0} & \textbf{71.3} & \textbf{92.7} & \textbf{75.8} & \textbf{76.2} & \textbf{70.1} & \textbf{77.4} \\
 \hline
   \multirow{2}{*}{b)} & 
    no matching & 76.2 & 68.5 & 91.2 & 72.8 & 73.6 & 66.3 & 74.8 \\ 
     &   no causal LM & 77.3 & 70.6 & 92.4 & 75.7 & 75.6 & 68.9 & 76.8 \\ \hline
 \end{tabular}

\end{table}

We conduct an ablation study to analyze the impacts of our proposed pretraining objectives: a) casual LM objectives at stage-1 unimodal pretraining on two generative tasks including code completion and math programming, b) text-code matching and causal LM objectives at stage-2 bimodal pretraining on an understanding task of text-to-code retrieval. 
We employ CodeT5+ 770M and  report the  results of three representative tasks over 10 datasets in \cref{tab:ablation}.
In CodeT5+, we found that  causal LM objective plays a crucial role in code completion and math programming tasks, observed by 
 a significant performance drop after removing it. This indicates causal LM can complement the span denoising objective and improve the generation capability of our models. 
Additionally, we found that the text-code matching objective is critical to the retrieval performance (a drop of $2.6$ avg. MRR over 6 datasets without it), implying this objective can learn a better bimodal representation that captures the fine-grained alignment between text and code.
Besides, we found that retrieval tasks can also benefit  from the joint training with causal LM objective despite their task differences.

\begin{table}[t]
\centering
\caption{
\textbf{Results of retrieval-augmented code generation:}
while other models are often used as either retrieval or generation models but not both, CodeT5+ can be easily adapted as both retriever and generator.
The model components are activated as an end-to-end retrieval-augmented code generation systems, leading to superior performance. 
EM: Exact Match, B4: BLEU-4, CB: CodeBLEU.
}
\label{tab:ra_gen}

\begin{tabular}{l|ccc|ccc}
\hline
\multirow{2}{*}{Model} & \multicolumn{3}{c}{Java} & \multicolumn{3}{|c}{Python} \\

                       & EM     & B4     & CB     & EM      & B4      & CB     \\
\hline
\multicolumn{7}{c}{Retrieval-based}                                     \\
\hline
BM25                   & 0.00   & 4.90   & 16.00  & 0.00    & 6.63    & 13.49  \\
SCODE-R 125M               & 0.00   & 25.34  & 26.68  & 0.00    & 22.75   & 23.92  \\
CodeT5+ 220M         & 0.00   & 28.74  & 31.00  & 0.00    & 27.30   & 26.51  \\
\hline
\multicolumn{7}{c}{Generative }    \\              
\hline
CodeBERT 125M              & 0.00   & 8.38   & 14.52  & 0.00    & 4.06    & 10.42  \\
GraphCodeBERT 125M          & 0.00   & 7.86   & 14.53  & 0.00    & 3.97    & 10.55  \\
PLBART 140M                & 0.00   & 10.10  & 14.96  & 0.00    & 4.89    & 12.01  \\
CodeT5+ 220M         & 0.00   & 10.33  & 20.54  & 0.00    & 4.40    & 13.88  \\
\hline
\multicolumn{7}{c}{Retrieval-Augmented Generative}              \\
\hline
REDCODER-EXT 125M+140M            & 10.21  & 28.98  & 33.18  & 9.61    & 24.43   & 30.21  \\
CodeT5+ 220M         & \textbf{11.66}  & \textbf{33.83}  & \textbf{40.60}  & \textbf{11.83}   & \textbf{31.14}   & \textbf{36.39}    \\
\hline
\end{tabular}
\end{table}

\begin{figure}[t]
	\centering
	\resizebox{\linewidth}{!} {
	\includegraphics{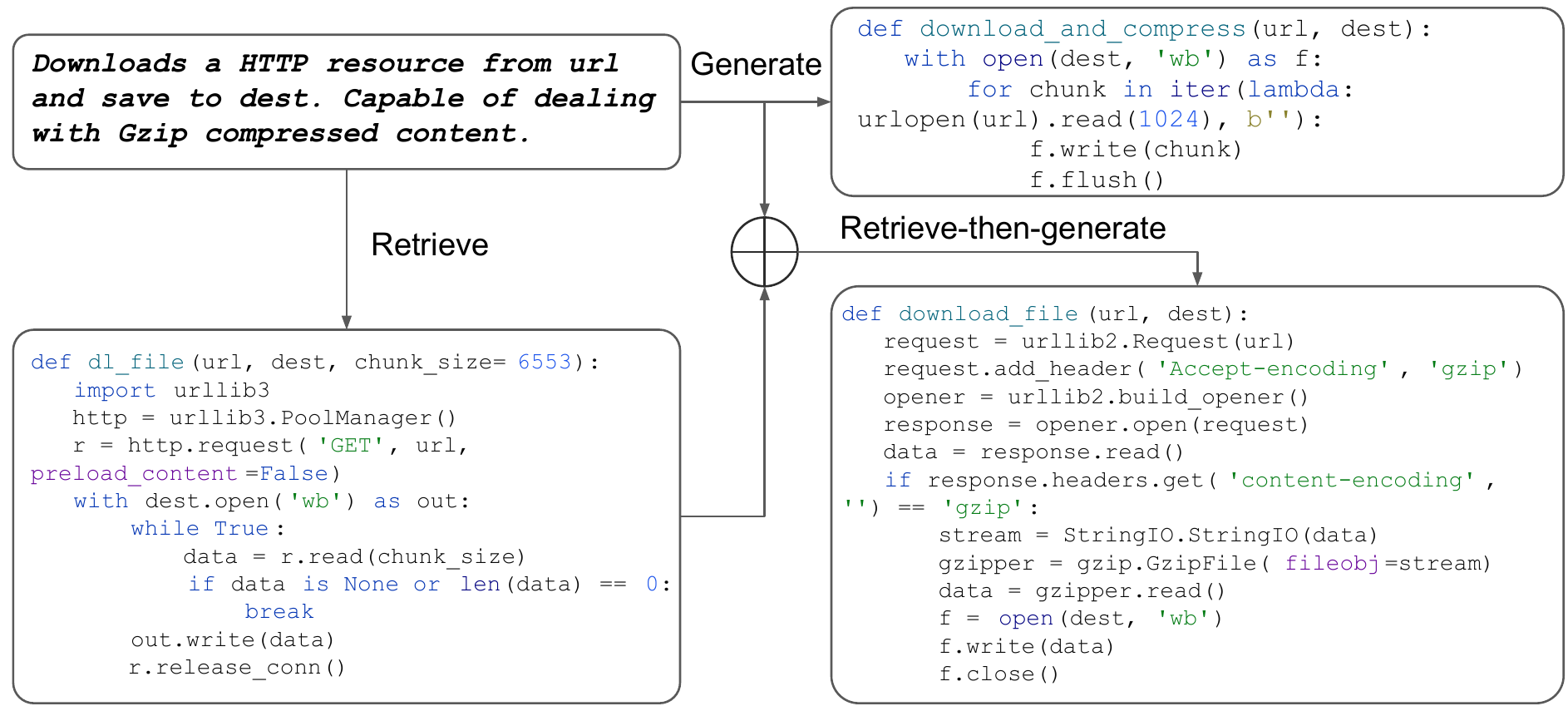}
	}
	\caption{
 \textbf{Example code generation output:}
 Our CodeT5+ retrieval-augmented generation model could retrieve relevant code context and use it to facilitate better code generation.}
	\label{fig:ra_code_gen_case}
\end{figure}

\subsection{Unified Retrieval-Augmented  Generation Paradigm}
\label{subsec:retrieval-gen}
As our model is capable of both code retrieval and generation, it can be naturally exploited as a unified semi-parametric retrieval-augmented generator. 
To explore this adaptation, we follow~\citet{ra_gen_sum} to evaluate two code generation tasks  by reversing the input and output order of code summarization on Java and Python and using their released deduplicated retrieval codebase.
We evaluate our models in three settings: retrieval-based, generative, and retrieval-augmented (RA) generative.  For the retrieval-based setting, we activate our encoder to retrieve the top-1 code sample as the prediction given a text query, while for the RA generative setting, we append the combination of top-$k$ retrieved samples ($k$=1 in our work) to the encoder input and activate the decoder.

As shown in \cref{tab:ra_gen},  we found that our CodeT5+ achieves  better results in all  categories, especially in the retrieval-based and RA generative setting. 
While the previous SoTA model REDCODER-EXT \citep{ra_gen_sum} separately employs GraphCodeBERT as the retriever and PLBART as the generator, our model can be flexibly used as an end-to-end system with both retrieval and generation capabilities.
We further include a qualitative case in \cref{fig:ra_code_gen_case}, where we found that the retrieved code provides crucial contexts (e.g., use ``urllib3'' for an HTTP request) to guide the generative process for more correct prediction.
In contrast, the generative-only model gives an incorrect prediction that only captures the concepts of ``download'' and ``compress''.
Additionally, we  analyze the effects of various top-$k$ retrievals on the code generation performance (see Appendix~\ref{app_ssec:topk}). 

\section{Conclusion}
We present CodeT5+, a new family of open code large language models with an encoder-decoder architecture that can flexibly operate in different modes (encoder-only, decoder-only, and encoder-decoder) to support a wide range of code understanding and generation tasks. To train CodeT5+, we introduce a mixture of pretraining tasks including span denoising, causal language modeling, contrastive learning, and text-code matching to learn rich representations from both unimodal code data and bimodal code-text data. Additionally,  we propose a simple yet effective compute-efficient training method to initialize our model with frozen off-the-shelf LLMs to efficiently scale up the model. We explore further instruction tuning to align the model with natural language instructions. 
Extensive experiments on a broad set of code intelligence  tasks over $20$ datasets have verified the superiority of our model. Particularly, on the zero-shot HumanEval code generation tasks, our instruction-tuned CodeT5+ 16B established new SoTA results of $35.0\%$ pass@1 and $54.5\%$ pass@10 against other open code LLMs and even surpasses the OpenAI code-cushman-001 model. Finally, we showcase the flexibility of CodeT5+ to  deploy as a unified retrieval-augmented generation system.

\bibliography{reference}
\bibliographystyle{abbrvnat}

% \newpage
\appendix
\section{Ethics Statement}

Advancements in code understanding and generation systems hold immense potential to create positive societal impacts by improving programming accessibility and enhancing developer productivity through natural language interfaces. However, deploying such systems at scale requires careful consideration of various ethical aspects, as extensively discussed by \citet{codex}.

One critical concern is the potential risk of generated code summaries or comments incorporating toxic or insensitive language, which can have detrimental effects. Several studies have explored techniques to address this issue, such as reinforcement learning \citep{ouyang2022training}, weighted decoding \citep{krause-etal-2021-gedi-generative} , and safety-specific control tokens  \citep{xu2020recipes}. These approaches aim to ensure non-toxic natural language generation, promoting responsible and ethical use of large language models for code.

Additionally, it is essential to recognize the broader intellectual property implications of code generation and retrieval systems before deployment. Deep learning models generating code may inadvertently introduce security vulnerabilities. To mitigate this risk, it is crucial to conduct expert reviews and rigorous security assessments before adopting such code. This review process ensures that the generated code meets necessary security standards, safeguarding against potential exploits and vulnerabilities.
In code retrieval scenarios, providing appropriate attribution to the source along with the retrieved results is paramount. This attribution not only respects the rights of code authors but also enhances transparency, traceability, and collaboration within the programming community. By acknowledging the original authors and promoting a collaborative, ethical, and legally compliant environment, code retrieval systems can foster knowledge sharing and contribute to a reputable programming ecosystem.

By considering these ethical considerations, we can promote the responsible deployment of large language models for code, maximizing their potential benefits while mitigating potential harms to individuals, communities, and the overall software ecosystem. It is imperative to prioritize safety, non-toxicity, intellectual property rights, security, and collaboration in the development and deployment of these systems, ensuring they align with ethical principles and societal needs.
 
\section{Bimodal Pretraining Details}
To expose the model on more diverse set of pretraining data, we employ a stage-wise pretraining process to first train CodeT5+ on large-scale code-only data with span denoising and causal language modeling (CLM) tasks, then train on smaller set of text-code bimodel data using text-code contrastive learning, matching, and causal LM tasks. Below, we provide detailed formulas for text-code contrastive learning and matching tasks at the second-stage pretraining on text-code pairs.

\paragraph{Text-Code Contrastive Learning} activates the encoder  to learn better unimodal (text/code) representations by computing a similarity score such that parallel text-code pairs have higher scores. Given a text T and a code C, we first learn representations $\mathbf{h}^t$ for text $T$ and  $\mathbf{h}^c$ for code $C$ by mapping the \texttt{[CLS]} embeddings to normalized lower-dimensional (256-d) representations from the encoder. Given a batch of $N$ text-code pairs, we obtain text vectors $\{\mathbf{h}^{t}\}_{i=1}^N$ and code vectors $\{\mathbf{h}^{c}\}_{i=1}^N$ to compute text-to-code and code-to-text and  similarities:
\begin{equation}
    s_{i,j}^{t2c} = \mathbf{h}^{t\top}_i \mathbf{h}^{c}_j, s_{i,j}^{c2t} = \mathbf{h}^{c\top}_i \mathbf{h}^{t}_j
\end{equation}
\begin{equation}
    p_i^{t2c}(T) = \frac{\exp{(s_{i,i}^{t2c}/\tau)}}{\sum_{j=1}^N \exp{(s_{i,j}^{t2c}/\tau)}}, p_i^{c2t}(C) = \frac{\exp{(s_{i,i}^{c2t}/\tau)}}{\sum_{j=1}^N \exp{(s_{i,j}^{c2t}/\tau)}} 
\end{equation}
where $s_{i,j}^{t2c}$ represents text-to-code similarity of text of $i$-th pair and code of $j$-th pair, and $s_{i,j}^{c2t}$ is the code-to-text similarity, $\tau$ is learned temperature parameter. $p_i^{t2c}(T)$ and $p_i^{c2t}(C)$ are the softmax-normalized text-to-code and code-to-text similarities for the $i$-th text and code. 

Let $\mathbf{y}^{t2c}(T)$ and $\mathbf{y}^{c2t}(C)$ denote the ground-truth one-hot similarity, where negative pairs have a probability of 0 and the positive pair has a probability of 1. The text-code contrastive loss from a  corpus $D$ of text-code pairs is defined as the cross-entropy H between $\mathbf{p}$ and $\mathbf{y}$:
\begin{equation}
    \mathcal{L}_{tcc} = \frac{1}{2} \mathbb{E}_{(T,C)\sim D}[H(\mathbf{y}^{t2c}(T), \mathbf{p}^{t2c}(T)) + H(\mathbf{y}^{c2t}(C), \mathbf{p}^{c2t}(C))]
\end{equation}

\paragraph{Text-Code Matching} activates the decoder with the bimodal matching functionality to predict whether a pair of text and code is positive (matched) or negative (unmatched). We employ the output embedding of the \texttt{[EOS]} token as the fused bimodal representation for a text-code pair ($T$, $C$), as this token attends to all the previous context for the text-code pair input. Followed by a linear layer and softmax, we compute a two-class probability $p^{tcm}(T,C)$ and define the matching loss:
\begin{equation}
    \mathcal{L}_{tcm} =  \mathbb{E}_{(T,C)\sim D} [H(\mathbf{y}^{tcm}(T,C), \mathbf{p}^{tcm}(T,C))]
\end{equation}
where $\mathbf{y}^{tcm}(T,C)$ is a 2-dimensional one-hot vector representing the ground-truth label.

\paragraph{Text-Code Causal LM.} This task focuses on a cross-modal causal LM objective between text and code through a dual multimodal conversion: text-to-code generation and code-to-text generation (i.e. code summarization). Let $\mathcal{L}_{t2c}$ and $\mathcal{L}_{c2t}$ denote the  losses for text-to-code and code-to-text generation. The full second-stage pretraining loss of our CodeT5+ is:
\begin{equation}
    \mathcal{L} = \mathcal{L}_{tcc}  + \mathcal{L}_{tcm} +  \mathcal{L}_{t2c} + \mathcal{L}_{c2t}
\end{equation}

\section{Additional Experimental Results}\label{app_sec:more_result}
In this section, we provide additional experimental results
including two understanding tasks  of code defect detection and clone detection from the CodeXGLUE benchmark~\citep{codexglue} (\cref{app_ssec: defect_clone}), analysis of the effects of top-k retrievals in retrieval-augmented code generation tasks (\cref{app_ssec:topk}), and more qualitative results in math programming tasks (\cref{app_ssec:math_qualitative}).

\subsection{Code Defect Detection and Clone Detection from CodeXGLUE}\label{app_ssec: defect_clone}
\begin{table}[t]
\centering
\caption{Results on two understanding tasks: code defect detection and code clone detection.}
\label{tab:defect_clone}

\begin{tabular}{l|c|ccc}
\hline
\multirow{2}{*}{Model}                & Defect & \multicolumn{3}{c}{Clone Detection} \\
\cline{2-5}
                & Acc    & Rec     & Prec    & F1      \\
\hline
% RoBERTa 125M        & 61.1   & 95.1       & 87.8         & 91.3    \\
CodeBERT 125M       & 62.1   & 94.7       & 93.4         & 94.1    \\
GraphCodeBERT 125M   & -      & 94.8       & 95.2         & 95.0    \\
UniXcoder 125M       & -      & 92.9       & \textbf{97.6}         & \textbf{95.2}    \\
CodeGen-multi 350M & 63.1 & 94.1 & 93.2 & 93.6  \\
\hline
PLBART 140M         & 63.2   & 94.8       & 92.5         & 93.6    \\
CodeT5 220M     & 65.8   & 95.1       & 94.9         & 95.0    \\

\hline
CodeT5+ 220M  & 66.1   & 96.4       & 94.1         & \textbf{95.2}    \\
CodeT5+ 770M & \textbf{66.7}   & \textbf{96.7}       & 93.5         & 95.1            \\
\hline
\end{tabular}
\end{table}

Defect detection is to predict whether a code is vulnerable to software systems or not, while clone detection aims to measure the similarity between two code snippets and predict whether they have a common functionality. 
We use benchmarks from CodeXGLUE~\citep{codexglue} and use  accuracy and F1 score as the metrics. 
In \cref{tab:defect_clone}, we can see CodeT5+ models achieve new SoTA accuracy of 66.7\% on the defect detection task.
For the clone detection task, our model achieves comparable results to SoTA models, where the performance increase tends to be saturated, observed by the close performance gaps between multiple baselines.

\subsection{Analysis on the Effects of Top-k Retrievals in Retrieval-augmented Code Generation}\label{app_ssec:topk}

\begin{table}[t]

\centering
\caption{Effects of varying top-$k$ retrievals in retrieval-augmented code generation tasks with our CodeT5+ 220M compared to the prior SOTA model of REDCODER-EXT that employs top-10 retrievals for augmentation.
EM: Exact Match, B4: BLEU-4, CB: CodeBLEU.
}
\label{tab:ra_gen_topk}

\begin{tabular}{l|ccc|ccc}
\hline
\multirow{2}{*}{Model} & \multicolumn{3}{c}{Java} & \multicolumn{3}{|c}{Python} \\
\cline{2-7}
                       & EM     & B4     & CB     & EM      & B4      & CB     \\
\hline
SOTA (top-10)         & 10.21  & 28.98  & 33.18  & 9.61    & 24.43   & 30.21  \\
\hline
Ours  &&&&&&      \\
top-1  & 11.66  & \textbf{33.83}  & 40.60  & 11.83   & 31.14   & 36.39    \\
top-2 & 11.57 & 33.26 & 40.74 & 11.78 & \textbf{31.21} & 36.58 \\
top-3 & 12.29 & 33.10 & 41.71 & 12.48 & 30.92 & 37.31 \\
top-4 & 12.42 & 32.08 & 41.94 & 12.73 & 30.40 & 37.60 \\
top-5 & \textbf{13.02} & 32.42 & \textbf{42.28} & \textbf{12.93} & 30.52 & \textbf{37.87} \\
top-10 & 12.86 & 31.38 & 42.24  & 12.84 & 29.79 & 37.79 \\
\hline
\end{tabular}

\end{table}

We further conduct an ablation study to analyze the effects of top-$k$ retrievals in retrieval-augmented code generation tasks and report the results in \cref{tab:ra_gen_topk} .
We found that increasing the number of retrievals can boost model performance which becomes saturated when $k$=5.
This saturation is due to the maximum sequence length of $600$, which might not be able to accommodate a large number of retrieved code samples. Overall, our CodeT5+ significantly outperforms the prior SOTA baseline which uses top-10 retrievals in all cases, even with only a top-1 retrieved code.

\subsection{Qualitative Results in Math Programming tasks}\label{app_ssec:math_qualitative}

\begin{figure*}[t]
	\centering
	\resizebox{\textwidth}{!} {
	\includegraphics{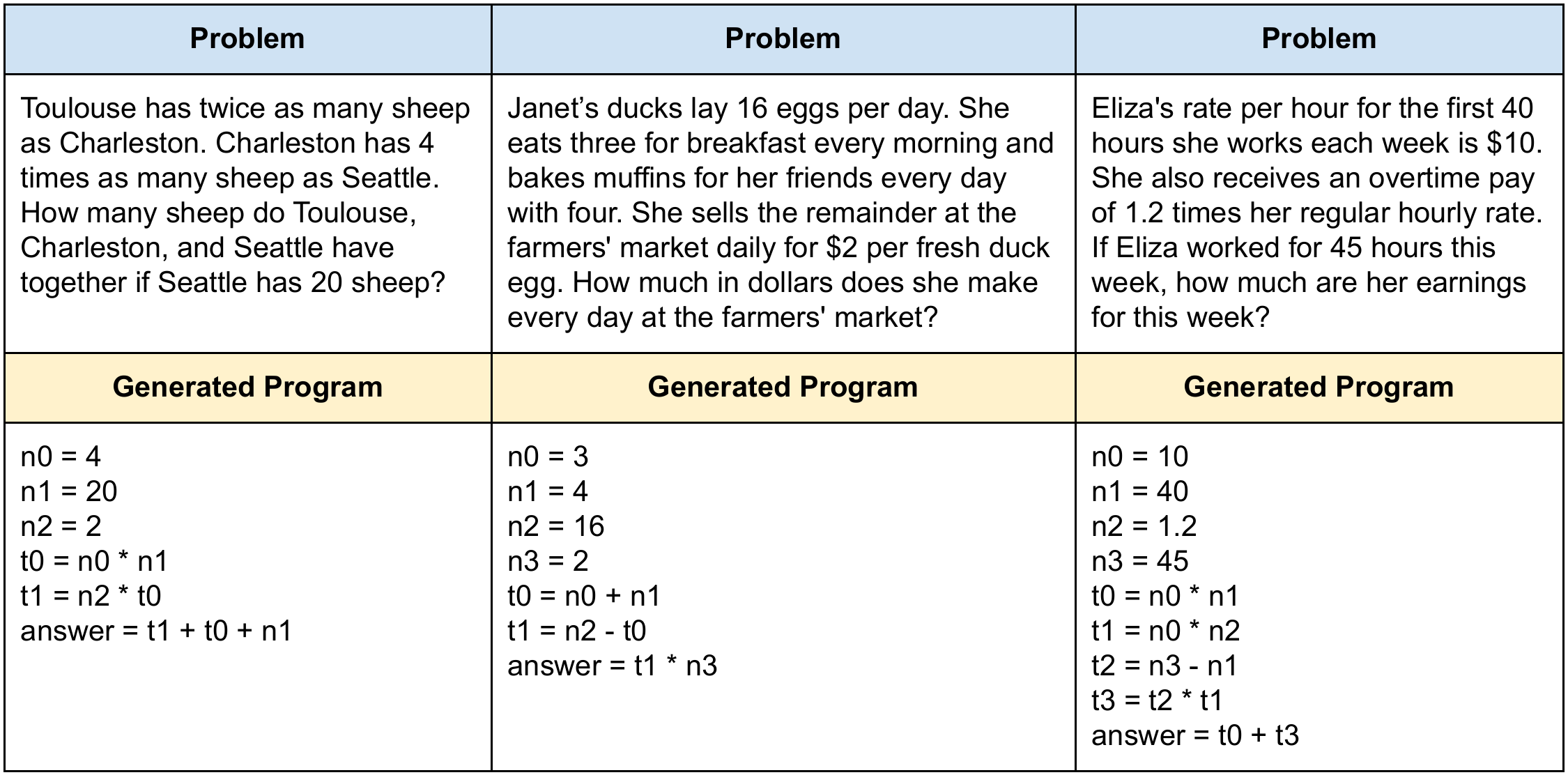}
	}
	\caption{Predictions of our model on GSM8K-Python}
	\label{fig:gsm_case}
\end{figure*}

\begin{figure*}[t]
	\centering
	\resizebox{\textwidth}{!} {
	\includegraphics{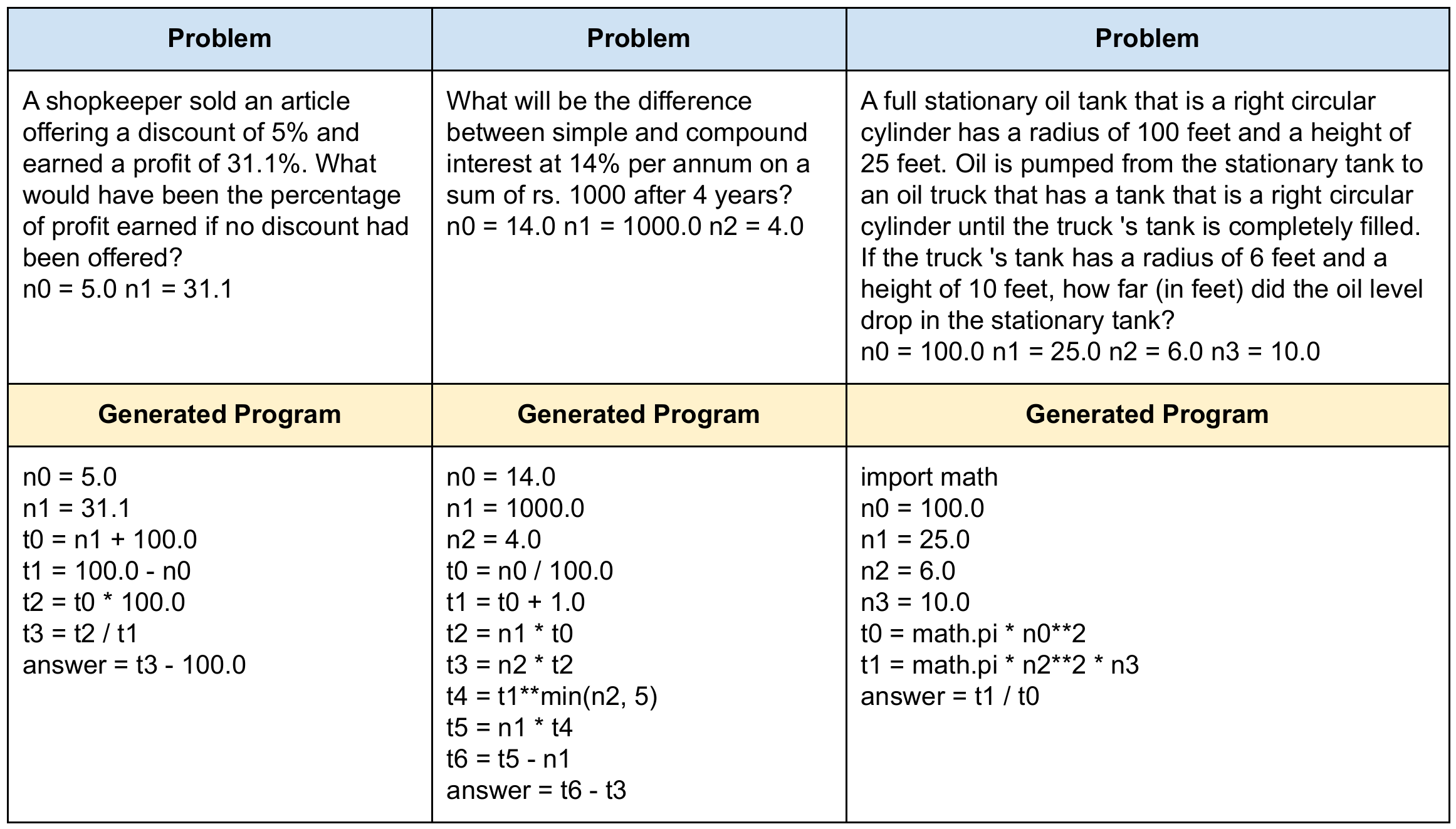}
	}
	\caption{Predictions of our model on MathQA-Python}
	\label{fig:mathqa_case}
\end{figure*}

For math programming tasks, we provide qualitative examples predicted by our models in \cref{fig:gsm_case} and \cref{fig:mathqa_case}.
Overall, we found CodeT5+ is able to generate decent programs that can solve the math problems in various levels of difficulties, i.e. from simple math operations to more complex problems with multiple reasoning steps. From the rightmost example of \cref{fig:mathqa_case}, we found that CodeT5+ is able to leverage some external libraries such as \emph{math} when synthesizing the solutions.

\section{Downstream Task Finetuning Details}\label{app_sec:finetune}

\subsection{Text-to-Code Retrieval}
 
 Text-to-code retrieval (or code search), is the task of finding the best code sample that is most relevant to a natural language query, from a collection of code candidates. 
 We experiment CodeT5+ with three major benchmarks: 
 CodeSearchNet (CSN) \citep{csn}, CosQA \citep{cosqa}, and AdvTest \citep{codexglue}. 
 CSN consists of six programming languages in total, and the dataset is curated by filtering low-quality queries through handcrafted rules, following \citep{graphcodebert}. 
 For instance, an example handcraft rule is to filter examples in which the number of tokens in query is shorter than $3$ or more than $256$. 
 % The resulting dataset statistics can be seen in Table \ref{tab:csn_data_stats}.
  % \input{tables/csn_data_stats}
 
 CosQA and AdvTest are two related benchmarks that are derived from the CSN data. 
 Specifically, instead of natural language queries, CosQA uses logs from Microsoft Bing search engine as queries, each of which is annotated by 3 human annotators \citep{cosqa}. 
 AdvTest is created from the Python split of the CSN data but the code samples are normalized with obfuscated variable names to better evaluate the understanding abilities of current models. 

 For training, we set the maximum sequence to 350 and 64 for code and text. We set the learning rate as 2e-5 and finetune the model for 10 epochs. We employ distributed training 
 % using Pytorch DistributedDataParallel\footnote{\url{https://pytorch.org/}} 
 on 8 A100s and the total batch size is 64. For momentum encoders, we maintain a separate text/code queue with a size of 57600, and allow the matching decoder to retrieve 64 hard negatives from the queues for hard negative mining.

\subsection{Code Summarization}

Code summarization is the task of generating a natural language summary of a code snippet. 
We use the task dataset from CodeXGLUE \citep{codexglue} which curated a code summarization benchmark from CSN data~\citep{csn}. 
The benchmark consists of six PLs: Ruby, JavaScript, Go, Python, Java, and PHP. It is the same clean version of CSN data that we use for text-to-code retrieval tasks.
For training, we set the maximum sequence length of the source
and target as 256 and 128, respectively. We use a learning rate of 2e-5, the batch size as 64 for 10 epochs of finetuning. We set the beam size as 5 in inference.

\subsection{Code Defect Detection}  

Defect detection is the task of classifying whether a code sample contains vulnerability points or not. 
We adopt the defect detection benchmark from CodeXGLUE \citep{codexglue} which curated data from the Devign dataset \citep{zhou2019devign}.
The dataset contains in total more than 27,000 annotated functions in C programming language. 
All samples are collected from popular open-source projects such as QEMU and FFmpeg. 
We follow \cite{codexglue} and adopt 80\%/10\%/10\% of the dataset as the training/validation/test split. 
For training, we set the learning rate as 2e-5, the batch size as 32, and the max sequence length as 512 to finetune the model for 10 epochs.

\subsection{Code Clone Detection}

The task of clone detection aims to detect whether any two code samples have the same functionality or semantics. 
We conduct experiments using the clone detection benchmark from CodeXGLUE \citep{codexglue}. 
The benchmark is curated from the BigClone Benchmark dataset \citep{svajlenko2014towards} and the resulting curated data consists of 901,724/416,328/416,328 examples for training/validation/test splits respectively. 
All samples are categorized into 10 different functionalities.
For finetuning, we set the learning rate as 2e-5 and finetune the model for 2 epochs. We set the
batch size as 10, and the max sequence length as 400.

\subsection{Code Completion}

In code completion,  given a source sequence containing a partial code sample, a model is required to generate the remaining part of the code sample. 
We conduct experiments on line-level code completion using two major benchmarks: 
PY150 \citep{py150} and  JavaCorpus \citep{javacorpus}. 
PY150 \citep{py150} consists of 150,000 Python source files collected from Github.
Among these samples, \cite{codexglue} selected 10,000 samples from different files from the test set of PY150 and then randomly sampled lines to be predicted for the code completion task.
The average numbers of tokens in the source sequence and target sequence are 489.1 and 6.6 respectively. 
JavaCorpus \citep{javacorpus} contains over 14,000 Java projects collected from GitHub.
Similarly to PY150, \citet{codexglue} selected 3,000 samples from different files from the test set of the dataset and randomly sampled lines to be predicted for the code completion task. 
The average numbers of tokens in the source and target sequence are 350.6 and 10.5 respectively. 

For both tasks, we set the learning rate as 2e-5 and batch size as 32, and set the maximum sequence length of 1024 for the decoder. We finetune the model for 30 epochs. During inference, we employ beam search with a beam size of 5.

\subsection{Math Programming}
Math Programming is the task of solving maths-based problems with programming. 
Compared to conventional code generation tasks, this task focuses more on computational reasoning skills.
The problem descriptions in this type of task are also more complex than conventional code generation tasks. 
We employ two major benchmarks for this task: MathQA-Python \citep{austin2021program} and GradeSchool-Math \citep{gsm}.

MathQA-Python \citep{austin2021program} is developed from the MathQA dataset \citep{amini-etal-2019-mathqa} where given a mathematical problem description in natural language, a system is required to solve this problem via generating a program that returns the final answer. 
\cite{austin2021program} translated these programs into Python programs and filtered for cleaner problems. 
In total, MathQA-Python contains  $\sim$24,000 problems, including 19,209/2,822/1,883 samples for training/validation/test splits. 

GradeSchool-Math \citep{gsm} (also known as GSM8K) has similar nature as MathQA. The benchmark focuses on problems with moderate difficulty that an average grade school student should be able to solve. 
In total, GSM data contains 8,500 problems, divided into 7,500 training and 1,000 testing problems. We translated the solution described in natural language  to Python programs by following  the construction process of MathQA-Python by~\citet{austin2021program}. Finally, we successfully converted 5,861 out of 7,500 training samples. 

For training, we set the maximum sequence length of the source
and target as 256 and 256 for MathQA-Python, and 246, 138 for GSM8k-Python. We use a learning rate of 2e-5 and a batch size of 32 for 30 epochs of finetuning.
During inference, we employ the beam size as 5 to get pass@1 results. For pass@80 and pass@100, we found they are quite sensitive to the diversity of the generation. We employ nucleus sampling with a temperature of $1.2$ and top-$p$=$0.95$.

\subsection{Retrieval-augmented Code Generation}
Developers often search for relevant code snippets from sources on the web such as GitHub or StackOverflow as references to aid their software development process. Motivated by this behaviour, we explore a retrieval-augmented code generation setting, where given a natural language description, a retriever first retrieves similar candidates in a search codebase and then augments the input for the generator to produce the target code.
Such retrieval-augmented generation (or retrieve-then-generate) paradigm has been widely used in open-domain question answering~\citep{dpr} in NLP and recently extended to some code-related tasks such as code generation and summarization~\citep{ra_gen_sum} with significant improvements.
As our CodeT5+ is capable of both retrieval and generation, it can be seamlessly adapted as a unified retrieval-augmented generator. This can bring unique benefits such as less computational cost compared to prior work that employs a different retriever and generator. We evaluate CodeT5+ on two Java and Python code generation datasets from the CodeXGLUE \cite{codexglue} benchmark following \citet{ra_gen_sum}. 

Specifically, we  leverage the encoder to encode the code snippet in the retrieval base and  build a search index with the \texttt{faiss} library \citep{johnson2019billion}. The search index is a set of representations (of $256$ dimensions) for all the code snippets  in the retrieval codebase.
Let $(x_i, y_i)$ denote one training instance where $x_i$ is the input text description and $y_i$ is the corresponding target code snippet.
we employ the same  encoder to obtain the embedding of $x_i$ and retrieve top-$k$ similar code samples from the search base using the L-2 similarity metric, with $k$ being a hyperparameter.  We ensure that the training example's target string ($y_i$) is not present in any of these $k$ retrieved samples.

After retrieving these top-$k$ relevant code samples, we combine them with a  special token \texttt{[SEP]} and concatenate it to the end of the source input $x_i$.
Unlike \cite{ra_gen_sum}, we do not augment docstrings or text descriptions and only  augment the code snippet for simplicity. 
We then finetune CodeT5+ on this augmented dataset. During inference, we retrieve similar code samples from the search base and augment these to input $x_i$.
For training, we set the maximum sequence length of the source
and target as 600 and 320. We use a learning rate of 2e-5, the batch size as 32 to finetune the model for 10 epochs. We set the beam size as 5 during inference with beam search.

\end{document}